\documentclass[english,preprint]{elsarticle}
\usepackage[T1]{fontenc}
\usepackage[latin9]{inputenc}
\usepackage{amsmath}
\usepackage{graphicx}
\usepackage{amssymb}
\usepackage{esint}
\usepackage{babel}
\usepackage {subfig}
\providecommand{\tabularnewline}{\\}

\numberwithin{equation}{section}

\begin{document}
\begin{frontmatter}
\title{Fast convolution based method for computing the signed distance
function and its derivatives: Linear solution to the non-linear eikonal problem}

\author[ufladdress]{Karthik S. Gurumoorthy\corref{mycorrespondingauthor}\fnref{currentaddr}}
\cortext[mycorrespondingauthor]{Corresponding author}
\fntext[currentaddr]{Present address: International Center for Theoretical Sciences, Tata Institute of Fundamental Research, TIFR Centre Building, Indian Institute of Science Campus, Bangalore, Karnataka, 560012, India. Ph:+91-80-23610109 (extn: 21)}
\ead{karthik.gurumoorthy@icts.res.in}

\author[ufl]{Anand Rangarajan}
\ead{anand@cise.ufl.edu}

\address[ufladdress]{Department of Computer and Information Science and Engineering,
University of Florida, Gainesville, Florida, USA}

\begin{abstract}
We present a fast convolution-based technique for computing an approximate,
signed Euclidean distance function $S$ on a set of 2D and 3D grid
locations. Instead of solving the non-linear, static Hamilton-Jacobi equation
($\|\nabla S\|=1$), our solution stems from first solving for a scalar field $\phi$
in a \emph{linear} differential equation and then deriving the solution for $S$
by taking the negative logarithm.
In other words, when $S$ and $\phi$ are related by $\phi = \exp \left(-\frac{S}{\tau} \right)$
and $\phi$ satisfies a specific linear differential equation corresponding to the extremum of a variational problem, we obtain the
approximate Euclidean distance function $S = -\tau \log(\phi)$ which converges to the true solution in the limit as $\tau \rightarrow 0$.
This is in sharp contrast to techniques like the fast marching and fast sweeping methods which
directly solve the Hamilton-Jacobi equation by the Godunov upwind discretization
scheme. Our linear formulation results in a closed-form
solution to the approximate Euclidean distance function expressible
as a discrete convolution, and hence efficiently
computable using the fast Fourier transform (FFT). Our solution also
circumvents the need for spatial discretization of the derivative
operator. As $\tau\rightarrow0$,
we show the convergence of our results to the true solution and also
bound the error for a given value of $\tau$. The differentiability
of our solution allows us to compute---using a set of convolutions---the
first and second derivatives of the approximate distance function.
In order to determine the sign of the distance function (defined to
be positive inside a closed region and negative outside), we compute
the winding number in 2D and the topological degree in 3D, whose computations can also be performed via
fast convolutions. We demonstrate the efficacy of our method through a set of experimental results.
\end{abstract}

\begin{keyword}
distance transform; convolution; fast Fourier transform; winding number; topological degree; Green's function
\MSC[2010] 65D18, 65M80
\end{keyword}
\end{frontmatter}

\section{Introduction\label{sec:Introduction}}
Euclidean distance functions (more popularly 
referred to as distance transforms) are widely used in image analysis and synthesis \cite{Osher02}.
The task here is to assign at each grid point a value
corresponding to the Euclidean distance to its nearest neighbor from
a given point-set. Formally stated: given a point-set $Y=\{Y_{k}\in\mathbb{R}^{D},k\in\{1,\ldots,K\}\}$
where $D$ is the dimensionality of the point-set and a set of equally
spaced Cartesian grid points $X$, the Euclidean distance function
problem requires us to assign 
\begin{equation}
\label{eq:Rdef}
R(X)=\min_{k}\|X-Y_{k}\|
\end{equation}
where $\|\cdot\|$  represents its Euclidean norm. In computational
geometry, this is the Voronoi problem \cite{Aggarwal87,deBerg08} and the solution
$R(X)$ can be visualized as a set of cones with the centers being
the point-set locations $\{Y_{k}\}_{k=1}^K$. The Euclidean distance function
problem is a special case of the eikonal equation where the forcing function
is identically equal to one and hence satisfies the differential equation  
\begin{equation}
\|\nabla R\|=1
\label{EucEikonal}
\end{equation}
everywhere, barring the point-set locations and the Voronoi boundaries where $R$ is not differentiable.
Here $\nabla R=(R_{x},R_{y})$ denotes the gradients of $R$. This is a nonlinear differential equation and an example of a static Hamilton-Jacobi equation.

Since the advent of the fast marching method \cite{Sethian96},
the literature is replete with pioneering works which have successfully
tackled this problem. Fast marching is an elegant technique which solves
for $R$ in $O(N\log N)$ time at the given $N$ grid locations
using the Godunov upwind discretization scheme. The $log N$ stems from the overhead of administering a priority queue data structure.  The ingenious work in \cite{Yatziv06} gives an $O(N)$ implementation of the fast marching method with a cleverly chosen \emph{untidy priority queue} data structure. 
Faster methods like the fast sweeping method \cite{Zhao05} employs Gauss-Seidel iterations
and finds the solution for $R$ in $O(N)$. Though it is computationally nicer and easier to implement than the fast marching method, the actual number of sweeps required for convergence depends on the problem at hand---heuristically $2^D$ sweeps are required in $D$ dimensions.
Fast sweeping methods have also been extended to the more general static Hamilton-Jacobi equation \cite{Kao02}
and also for the eikonal equation on non-regular grids \cite{Qian07,Kao08}. A Hamiltonian approach to solve the eikonal equation can be found
in \cite{Siddiqi99}. Methods based on geometric consideration of the reverse Huygens's principle can be seen in \cite{Tsai02}.
In \cite{Aggarwal87}, the distance function is extracted by constructing the Voronoi diagram that is shown to be optimally linear in the number of data points. Other sweeping based methods for constructing the Euclidean distance function includes the work in \cite{Danielsson80}.

In this article we provide a detailed description and extend our previous
work on computing the signed Euclidean distance functions from point-sets
\cite{Gurumoorthy09,Rangarajan09}. Since our approach uses the standard discrete Fourier transform (DFT) formulation, we compute the distance function only from a discrete set of source points positioned in a regular Cartesian grid. Moving away from the DFT setting (and as a result sacrificing the speed of computation), extension of this algorithm to curves can be found in \cite{Sethi12}.
As we motivate our method from a variational perspective---which we consider
to be simultaneously novel and illuminating compared to our previous formulations---the proofs involved here
are quite different. The intriguing aspect of our approach is that the nonlinear Hamilton-Jacobi equation (\ref{EucEikonal}) is obtained in
the limit as $\tau\rightarrow0$ of a linear differential equation. Let $S$ denote the
approximate Euclidean distance function (with the nature of the approximation made clear below). When we express $S$ as the exponent of
a scalar field $\phi$, specifically $\phi(X)=\exp\left(-\frac{S(X)}{\tau}\right)$,
and if $\phi(X)$ is the solution to a specific variational problem satisfying its corresponding linear Euler-Lagrange equation,
we show that $S(X)$ converges to the true Euclidean distance function $R(X)$ and its gradient magnitude ($\|\nabla S\|$)
also satisfies (\ref{EucEikonal}) as $\tau \rightarrow 0$.
Consequently, instead of solving the non-linear Hamilton-Jacobi equation we solve for the function $\phi$ (taking advantage of its linearity),
and then compute an approximate distance function from its exponent for small values of $\tau$.
This computational procedure would be approximately equivalent to
solving the original Hamilton-Jacobi equation. Our linear approach results in a closed-form solution which can be expressed as a discrete
convolution and computed in $O(N\log N)$ time using a fast Fourier transform (FFT) \cite{Bracewell99} where $N$ is the number of grid
points. The major advantage of our method is that the closed-form solution circumvents the need
for spatial discretization of the derivative operator in (\ref{EucEikonal}),
a problem that permeates the Hamilton-Jacobi solvers \cite{Sethian96, Zhao05}.
This accounts for improved accuracy of our technique as demonstrated in
Section~\ref{sec:Experiments}. However, a minor caveat of our method is that
the resultant Euclidean distance function is an approximation
since it is obtained for a small but non-zero value of
$\tau$, but nevertheless converges to the true solution as $\tau\rightarrow0$. 

The linear approach gives only an unsigned distance function.
We complement this by independently finding the sign of the distance
function in $O(N\log N)$ time on a regular grid in 2D and 3D.
We achieve this by obtaining the \emph{winding number}
for each location in the 2D grid and its equivalent concept, the
\emph{topological degree} in 3D. We show that just as in the case
of the unsigned Euclidean distance function, the winding number and the topological
degree computations can be written in closed-form, expressed
as discrete convolutions and efficiently computed using FFTs. We are
not aware of any previous work that uses the winding number
and topological degree approaches to compute signed distance functions. Furthermore, it is not easy
to obtain the gradient of the signed distance function via the Hamilton-Jacobi solvers due to the lack of differentiability
of their solution. Since our method results in a differentiable closed-form
solution, we can leverage it to determine these gradient quantities. As
before, the gradients can also be expressed in the form of discrete
convolutions and computed using FFT's. The gradients themselves converge to
their true values as $\tau \rightarrow 0$. To our knowledge fast computation of the derivatives of the distance function
on a regular grid using discrete convolutions is new. 

The paper is organized as follows.  In Section~\ref{sec:waveequation} we derive the
linear equation formalism for the Euclidean distance function problem and 
show convergence of our solution to the true solution
as $\tau\rightarrow0$. We provide an approximate closed-form solution to compute the distance function
and give an error bound between the
computed and true distance functions for a given value of $\tau$.
Section~\ref{sec:EfficientComp} explains how the closed-form solution can be represented in the form of a discrete convolution
and computed using fast Fourier transforms. In Sections~\ref{sec:SignedDist}
and \ref{sec:Fastderivatives} we compute the winding number (in
2D), the topological degree (in 3D) and the derivatives of the
distance function by again expressing these quantities as discrete
convolutions. In Section~\ref{sec:Experiments} we provide anecdotal evidence for the usefulness of our method by furnishing
both experimental results and comparisons to standard techniques. We finally conclude in Section~\ref{sec:Conclusion}.

\section{Linear differential equation approach for Euclidean distance functions}
\label{sec:waveequation}
Recall that our objective is to compute the Euclidean distance function $R$
defined in (\ref{eq:Rdef}), at a set of grid locations $X$ from the
point-set $\{Y_k\}_{k=1}^K$. To this end, consider the following variational
problem for a function $\phi(X)$ namely, 
\begin{equation}
\label{eq:varProblem}
I\left[\phi\right] = \tau^2 \int_{\Omega} \|\nabla \phi\|^2 dX + \int_{\Omega} |\phi-\psi^{\tau}|^2 dX,
\end{equation}
where $\tau$ is a free parameter  independent of $X$ and $\Omega$ is the domain of integration. Here,
$\psi^{\tau}(X)$---a function whose definition depends on $\tau$---represents
the initial scalar field concentrated around the source locations
$\{Y_k\}_{k=1}^K$ in the limit as $\tau \rightarrow 0$. We define
$\psi^{\tau}(X)$ as 
\begin{equation}
\label{eq:phi0}
\psi^{\tau}(X) \equiv \sum_{k=1}^K \psi_k^{\tau}(X),
\end{equation}
where $\psi_k^{\tau}(X)$ is chosen such that it is \emph{square integrable}
with its support sequentially converging towards the point source $Y_k$ as
$\tau$ approaches zero and asymptotically behaves like the square-root of a
$\delta$ function centered around $Y_k$. The square integrability (to one)
constraint changes its functional form in accordance with the spatial
dimension, as explicated in the subsequent sections. 

The Euler-Lagrange equation corresponding to the extremum of $I[\phi]$
computed over the scalar field $\phi$ is given by the \emph{linear} equation 
\begin{equation}
\label{eq:phiEquationwithtau}
-\tau^2 \nabla^2 \phi + \phi = \psi^\tau,
\end{equation}
where $\nabla^2$ stands for the Laplacian operator. We may be tempted to
replace $\psi^\tau$ in (\ref{eq:phiEquationwithtau}) with a combination of
delta functions each centered around $Y_k$ and obtain an \emph{inhomogeneous
  screened Poisson} partial differential equation. But as the delta functions
are not square-integrable, they cannot be incorporated into the variational
framework given in (\ref{eq:varProblem}). Defining $\psi^\tau$ as in
(\ref{eq:phi0}) forces it to behave like the square-root of a $\delta$
function as $\tau \rightarrow 0$ and hence is square-integrable for all values
of $\tau$. 

Armed with the above set-up we realize that when we relate
$\phi(X)\equiv\exp\left(-\frac{S(X)}{\tau}\right)$ and $\phi$ satisfies
(\ref{eq:phiEquationwithtau}) $S$ asymptotically converges to the true
solution $R$ in the limit as $\tau\rightarrow0$ as elucidated in
Section~\ref{sec:convergenceproof}. Furthermore, even the functions in $\nabla S$
namely ($S_x,S_y$) converge to their true values ($R_x,R_y$) and hence the
gradient magnitude $\|\nabla S \|$ satisfies the Hamilton-Jacobi
equation (~\ref{EucEikonal}) as $\tau\rightarrow0$. We prove the convergence
analysis for the gradient functions in Section~\ref{sec:derivConvergence}.  
This relationship motivates us to solve for the linear
equation in (\ref{eq:phiEquationwithtau}) instead of the non-linear 
eikonal equation and then compute the distance function via  
\begin{equation}
S(X)=-\tau\log\phi(X).
\label{SfromPhi}
\end{equation}

\subsection{Solution for the Euclidean distance function}
\label{sec:ClosedFormSol}
We now derive the solution for $\phi(X)$ (in 1D, 2D
and 3D) which satisfies (\ref{eq:phiEquationwithtau}) and then for $S(X)$
using the relation in (\ref{SfromPhi}). Since it is meaningful to assume that $S(X)$ approaches infinity
for points at infinity, we can use Dirichlet boundary conditions $\phi(X)=0$
at the boundary of an unbounded domain. The validity of our variational formalism even for an unbounded domain can be seen from \cite{Wahba90}.
Note that the variational approach only provides a mechanism to obtain the linear differential equation for $\phi$ in (\ref{eq:phiEquationwithtau})---which is the main focus of our paper.
The functional in (\ref{eq:varProblem}) provides orienting intuition for the origin of the differential equation---hence the lack of a fully formal treatment.
Using a Green's function approach
\cite{Abramowitz64}, we can write expressions for the solution $\phi$.
The form of the solution for $G$ \cite{Abramowitz64} in 1D, 2D
and 3D is given by\footnote{The representation of Green's function involves two variables $X$ and $Y$ and its value depends on the difference $X-Y$. For brevity we express $G(X,0)$ as $G(X)$ in this paper.}

\noindent \textbf{1D:} 
\begin{equation}
G(X)=\frac{1}{2\tau}\exp\left(-\frac{|X|}{\tau}\right),
\label{GreensFunction1D}
\end{equation}

\noindent \textbf{2D:} 
\begin{equation}
G(X)  =  \frac{1}{2\pi\tau^{2}}K_{0}\left(\frac{\|X\|}{\tau}\right)
\label{GreensFunction2D} \approx  \frac{\exp\left(-\frac{\|X\|}{\tau}\right)}{2\tau\sqrt{2\pi\tau\|X\|}},\,\frac{\|X\|}{\tau}\gg0.25 
 \end{equation}
where $K_{0}$ is the modified Bessel function of the second kind,

\noindent \textbf{3D:} 
\begin{equation}
G(X)=\frac{1}{4\pi\tau^{2}}\frac{\exp\left(-\frac{\|X\|}{\tau}\right)}{\|X\|}.
\label{GreensFunction3D}
\end{equation}
The solution for $\phi$ can then be obtained via convolution as 
\begin{equation}
\phi(X)= G(X)\ast\psi^{\tau}(X) = \sum_{k=1}^K G(X) \ast \psi_k^{\tau}(X).
\label{eq:phiSolution}
\end{equation}
$S(X)$ can then be recovered from its exponent using the relation~\ref{SfromPhi}.

\subsection{Proofs of convergence}
\label{sec:convergenceproof}
Since we require $\psi_k^{\tau}(X) $ to be square integrable to one and behave
like a square-root of the $\delta$ function centered at $Y_k$ as $\tau
\rightarrow 0$, its definition varies with the spatial dimension $D$ and is
given by:\\ 
\noindent \textbf{1D:} 
\begin{equation}
\psi_k^{\tau}(X) = \left \{ 
                              \begin{array} {ll}
                             \frac{1}{\sqrt{\tau}} & \mbox{for  } Y_k-\frac{\tau}{2} \leq X \leq Y_k+\frac{\tau}{2}; \\
                             0 & \mbox{otherwise}.
                             \end{array} 
                             \right .
\end{equation}
\noindent \textbf{2D:}
\begin{equation}
\psi_k^{\tau}(X) = \left \{ 
                              \begin{array} {ll}
                             \frac{1}{\tau}; & x_k-\frac{\tau}{2} \leq x \leq x_k+\frac{\tau}{2}, \\
                             &  y_k-\frac{\tau}{2} \leq y \leq y_k+\frac{\tau}{2}; \\
                             0 & \mbox{otherwise}.
                             \end{array} 
                             \right .
\end{equation}
\noindent \textbf{3D:} 
\begin{equation}
\psi_k^{\tau}(X) = \left \{ 
                              \begin{array} {ll}
                             \frac{1}{\tau^{\frac{3}{2}}}; & x_k-\frac{\tau}{2} \leq x \leq x_k+\frac{\tau}{2}, \\
                             &  y_k-\frac{\tau}{2} \leq y \leq y_k+\frac{\tau}{2},\\
                             &  z_k-\frac{\tau}{2} \leq z \leq z_k+\frac{\tau}{2},\\
                             0 & \mbox{otherwise}.
                             \end{array} 
                             \right .
\end{equation}
In 2D and in 3D the grid location $X$ and the point source $Y_k$ are
represented by $X=(x,y)$ and $Y_k=(x_k,y_k)$ and 
$X=(x,y,z)$ and $Y_k=(x_k,y_k,z_k)$ respectively. The proofs of convergence of
$S(X)$---obtained from the exponent of $\phi(X)$---to the 
true solution $R(X)$ given in (\ref{eq:Rdef}) as $\tau\rightarrow0$ are
relegated to \ref{sec:proof}. 

\subsection{Closed-form expression} 
A careful study of the proof in \ref{sec:proof} where we show convergence of $S(X)$ to the true
solution $R(X)$ encourages us to approximate the integral $\int_{\mathcal{B}_k^{\tau}(X)} G(Z)
dZ$---obtained as a result of convolving the Green's function $G$ with
$\psi_k^{\tau}$---with $\tau^D G(X-Y_k)$ at small values of $\tau$. Here $D$ is the
spatial dimension. This approximation is cogent for the following reasons:   
\begin{itemize}
\item First,  the integral region $\mathcal{B}_k^{\tau}(X)$ is considered
  only around $X-Y_k$ with its area/volume dwindling to zero as $\tau$
  approaches zero (please refer to equations (\ref{eq:IntergralRegion1D}) and
  (\ref{eq:IntergralRegion2D})). Hence we can replace the integral
  $\int_{\mathcal{B}_k^{\tau}(X)} G(Z) dZ$ by its Reimann summation 
  assuming that $G(Z)$ is \emph{constant} over $\mathcal{B}_k^{\tau}(X)$ and
  equals its mid-point value $G(X-Y_k)$.  
\item Second, our proof technique given in Section~\ref{sec:convergenceproof}
  attests to the fact that substituting the integral $\int_{\mathcal{B}_k^{\tau}(X)} G(Z)
  dZ$ by either $\tau^D \sup_{Z \in \mathcal{B}_k^{\tau}(X)} G(Z) $ or
  $\tau^D  \inf_ {Z \in \mathcal{B}_k^{\tau}(X)} G(Z)$  still allows us to establish
  convergence to the true solution. Since 
\begin{equation}
\inf_ {Z \in \mathcal{B}_k^{\tau}(X)} G(Z) \leq G(X-Y_k) \leq \sup_{Z \in \mathcal{B}_k^{\tau}(X)} G(Z), 
\end{equation}
our approximation is sound for small values of $\tau$. 
\item Third, it provides us with a \emph{closed-form} solution for $\phi$ as described below.
\end{itemize}
Hence we approximate the solution for the scalar field $\phi$ as
\begin{equation}
\label{eq:phiapproxSol}
\phi(X) \approx  \tau^{-\gamma} \tau^D \sum_{k=1}^{K}G(X-Y_k),
\end{equation}
at small values of $\tau$. Here $\tau^{-\gamma}$ equals
$\psi_k^{\tau}(Y_k)$. The value of $\gamma$ depends on the spatial dimension
$D$ as explained in Section~\ref{sec:convergenceproof}. 

Based on the nature of the expression for the Green's function $G$, it
is worth highlighting the following \emph{very} important point. The Green's
function in either 1D, 2D or 3D takes the form 
\begin{equation}
\lim_{\tau\rightarrow0}\frac{\exp\left\{ -\frac{\|X\|}{\tau}\right\} }{c\tau^{d}\|X\|^{p}}=0,\,\mathrm{for}\,\|X\|\neq0\label{eq:gfbogacity}
\end{equation}
for $c,d$ and $p$ being constants greater than zero. In the limiting case of $\tau\rightarrow0$, if we define 
\begin{equation}
\tilde{G}(X)\equiv C \exp\left(-\frac{\|X\|}{\tau}\right)
\label{GTildeDefinition}
\end{equation}
 for some constant $C$, then $\lim_{\tau\rightarrow0}|G(X)-\tilde{G}(X)|=0,\,\mathrm{for}\,\|X\|\neq0$
and furthermore, the convergence is \emph{uniform} for $\|X\|$ away
from zero. Therefore, $\tilde{G}(X)$ provides a very good approximation
for the actual unbounded domain Green's function as $\tau\rightarrow0$. For a fixed value of $\tau$ and $X$,
the difference between the Green's functions is $O\left(\frac{\exp\left(-\frac{\|X\|}{\tau} \right )}{\tau^2} \right)$
which is relatively insignificant for small values of $\tau$ and for all $X
\neq 0$.
Moreover, using $\tilde{G}$ also avoids the singularity at the origin in the 2D
and 3D cases. The above observation motivates us to compute the solution
for $\phi$ by convolving with $\tilde{G}$---instead of the actual
Green's function $G$ as in (\ref{eq:phiapproxSol})---to get
\begin{equation}
\phi(X) \approx  \tau^{-\gamma} \tau^D C  \sum_{k=1}^{K}\exp\left(-\frac{\|X-Y_{k}\|}{\tau}\right).
\end{equation}
The approximate Euclidean distance function computed based on the relation in (\ref{SfromPhi}) is then given by
\begin{equation}
S(X)\approx (\gamma-D) \tau \log \tau -\tau \log C - \tau \log \left\{ \sum_{k=1}^{K}\exp\left(-\frac{\|X-Y_{k}\|}{\tau}\right)\right\}.
\end{equation}
Since $(\gamma-D) \tau \log \tau$ and $\tau \log C$ are constants independent
of $X$ and converge to zero as $\tau \rightarrow 0$, they can be ignored while
solving for $S$ at small values of $\tau$. Hence the scalar field $\phi (X)$
can be further approximated as
\begin{equation} 
\label{eq:wavefunc}
\phi(X) \approx \sum_{k=1}^{K} \exp\left(-\frac{\|X-Y_k\|}{\tau}\right)
\end{equation} 
and the approximate Euclidean distance function equals
\begin{equation}
S(X) = -\tau  \log \left\{ \sum_{k=1}^{K}\exp\left(-\frac{\|X-Y_{k}\|}{\tau}\right)\right\}.
\label{STilde}
\end{equation}
We would like to underscore that the approximate solution for $\phi(X)$ given
in (\ref{eq:wavefunc}) possesses many desirable properties. 
\begin{itemize}
\item Notice that as $\tau \rightarrow 0$ $\phi(Y_k) \rightarrow 1$ at the given point-set
locations $\{Y_k\}_{k=1}^K$. Using the relation in (\ref{SfromPhi}) we get $S(Y_k)
\rightarrow 0$ as $\tau \rightarrow 0$ satisfying the initial
conditions.
\item For small values of $\tau$ the quantity
$\sum_{k=1}^{K}\exp\left(-\frac{\|X-Y_{k}\|}{\tau}\right)$ can be replaced 
by $\exp\left(-\frac{R(X)}{\tau}\right)$ where
$R(X)=\min_{k}\|X-Y_{k}\|$---the true Euclidean distance function. Hence we
get 
\begin{equation}
S(X) \approx-\tau \log\exp\left(-\frac{R(X)}{\tau}\right)=R(X). 
\end{equation}
\item $\phi(X)$ can be efficiently computed using the fast Fourier transform
as discussed in Section~\ref{sec:EfficientComp}. 
\end{itemize}
Hence for all practical, computational purposes we consider the expression given in
(\ref{eq:wavefunc}) as the solution for $\phi(X)$. The bound derived below
between $S(X)$ and $R(X)$ highlights the proximity between the computed and
actual Euclidean distance functions. 

\subsection{Error bound between the obtained and
  true Euclidean distance function}
\label{sec:ErrorBound}
Let $k_0$ denote the index of the source-point closest to $X$, i.e, $R(X)
=\|X-Y_{k_0}\|$. If $X$ lies on the Voronoi boundary, we may choose any one of
the multiple closest source-points and label it as $Y_{k_0}$. Multiplying and
dividing the expression for $\phi(X)$ in (\ref{eq:wavefunc}) by $\exp\left(
\frac{R(X)}{\tau}\right)$ we have 
\begin{equation}
S(X)  = R(X)-\tau  \log \left\{ \sum_{k=1,k \neq
  k_0}^{K}\exp\left(-\frac{\|X-Y_{k}\|-R(X)}{\tau}\right)\right\} \leq R(X). 
\label{Bound1}
\end{equation}
Since $\|X-Y_k\| \geq R(X)$, the term
$\exp\left(-\frac{\|X-Y_{k}\|-R(X)}{\tau}\right)$ equals one independent of
$\tau$ when $\|X-Y_k\| = R(X)$ or strictly decreases and converges to zero as
$\tau \rightarrow 0$ when $\|X-Y_k\| > R(X)$. In either case, $S(X)$
\emph{strictly increases} and converges to $R(X)$ in the limit as $\tau
\rightarrow 0$. 

 Moreover, we also get the inequality
 \begin{equation}
S(X) \geq  -\tau\log\left \{ K\exp\left(-\frac{R(X)}{\tau}\right)\right \} = -\tau\log K+R(X),
 \end{equation}
and along with (\ref{Bound1}) gives the bound 
\begin{equation}
\label{eq:errorBound}
|R(X)-S(X)|\leq\tau\log K
\end{equation}
which is very tight as: (i) it scales only as the logarithm of the cardinality of
the point-set ($K$) and (ii) can be made arbitrarily small by
choosing a small but non-zero value of $\tau$.

\section{Efficient computation of the approximate unsigned Euclidean distance
function}
\label{sec:EfficientComp}

We now provide a fast $O(N\log N)$ convolution-based method to compute
an approximate distance transform on a set of $N$ grid locations $\left\{ X_{i} \right \}_{i=1}^N$. 
The solution for $\phi(X)$ given in (\ref{eq:wavefunc}) at the $N$ grid locations can be represented as the
\emph{discrete convolution} between the functions 
\begin{equation}
f(X)\equiv\exp\left(-\frac{\|X\|}{\tau}\right)
\label{eq:function_f}
\end{equation}
 computed at the grid locations, with the function $g(X)$ which takes
the value $1$ at the point-set locations and $0$ at other grid locations, i.e, 
\begin{equation}
g(X) \equiv \sum_{k=1}^K \delta_{\mathrm{kron}}(X-Y_k) 
\label{eq:function_g}
\end{equation}
where,
\begin{equation}
\delta_{\mathrm{kron}}(X-Y_k) \equiv \left\{ 
	\begin{array}{ll}
         1 & \mbox{if $X = Y_k$};\\
        0 & \mbox{otherwise}.
        \end{array} 
        \right. 
 \label{eq:krondelta}        
\end{equation}
We would like to press the following point home. The Kronecker delta function
$\delta_{\mathrm{kron}}(X-Y_k) $ used in the definition of  $g(X)$ in (\ref{eq:function_g}) is \emph{not} a replacement for the function
$\psi_k^{\tau}(X)$ used to define $\psi^{\tau}(X)$ in (\ref{eq:phi0}). We
define $g(X)$ to be a sum of Kronecker delta functions so that the solution
for $\phi(X)$ given in (\ref{eq:wavefunc}) at the $N$ grid locations can be
written in the form of a discrete convolution. 

By the convolution theorem \cite{Bracewell99}, a discrete convolution
can be obtained as the inverse Fourier transform of the product of
two individual transforms, which for two $O(N)$ sequences can be
computed in $O(N\log N)$ time \cite{Cooley65}. One just
needs to compute the discrete Fourier transform (DFT) of the sampled
version of the functions $f(X)$ and $g(X)$, compute their point-wise
product and then compute the inverse discrete Fourier transform. Taking
the logarithm of the inverse discrete Fourier transform and multiplying
it by $(-\tau)$ gives the approximate Euclidean distance function.
The algorithm is spelled out in Table~\ref{tab:ApproxEuclid}.

\begin{table}
\begin{centering}
\caption{Approximate Euclidean distance function algorithm\label{tab:ApproxEuclid}}
\par\end{centering}
\centering{}\begin{tabular}{cl}
\hline 
1.  & Compute the function $f(X)=\exp\left(-\frac{\|X\|}{\tau}\right)$
at the grid locations.\tabularnewline
2.  & Define the function $g(X)$ which takes the value $1$ at the point-set
locations \tabularnewline
 & and $0$ at other grid locations.\tabularnewline
3.  & Compute the FFT of $f$ and $g$, namely $F_{\mathrm{FFT}}(U)$ and
$G_{\mathrm{FFT}}(U)$ respectively.\tabularnewline
4.  & Compute the function $H(U)=F_{\mathrm{FFT}}(U)G_{\mathrm{FFT}}(U)$.\tabularnewline
5.  & Compute the inverse FFT of $H(U)$ to obtain $\phi(X)$. \tabularnewline
6.  & Take the logarithm of $\phi(X)$ and multiply it by $(-\tau)$ to
recover \tabularnewline
 & the approximate Euclidean distance function.\tabularnewline
\hline
\end{tabular}
\end{table}

\subsection{Computation of the approximate Euclidean distance function in higher
dimensions}
Our technique has a straightforward generalization to higher dimensions.
Regardless of the spatial dimension, the approximate Euclidean distance
function ($S$) can be computed by exactly following the steps
delineated in Table~\ref{tab:ApproxEuclid}. It is worth accentuating
that computing the discrete Fourier transform using the FFT is
always $O(N\log N)$ \emph{irrespective} of the spatial
dimension\footnote{Even though the actual number of grid points ($N$) increases with dimension the solution is always $O(N\log N)$ in the number of grid points.}. Hence, in all dimensions $S$ can be computed at the given
$N$ grid points in $O(N\log N)$. This speaks for the scalability
of our technique which is generally not the case with other methods like KD-Trees \cite{deBerg08}.

\subsection{Numerical issues}
\label{sec:numericalissues}
In principle we should be able to run our technique at a very small
value of $\tau$ and obtain impressive results. However, for values of $\tau$ very close to zero, $f(X)$ tapers off very quickly and hence for grid locations which are far away from the point-set the convolution performed via FFT using a double precision-based implementation tends to be erroneous.
To this end, we turned to the GNU MPFR multiple-precision arithmetic library which
provides arbitrary precision arithmetic with correct rounding \cite{Fousse07}. 
MPFR is based on the GNU multiple-precision library (GMP) \cite{GMP}. It enables us to run our technique at low values of $\tau$. Advanpix \cite{Advanpix} is another $\mbox{MATLAB}^{\textregistered}$ software that supports arbitrary precision. 

\subsection{Is the aforesaid issue a major pitfall?}
It is natural for a reader to feel that our proposed method suffers from a fundamental impediment of requiring high precision machinery to serve its purpose. While the concern seems genuine, we would like to underscore that our technique can still be implemented in 64-bit precision (double data type) at a slightly higher value of $\tau$. In the last 4 decades we have seen rapid growth from 8-bit microprocessors (for e.g 8008 microprocessor launched by Intel in 1972) to the current 64-bit computing paradigm based on registers, address buses or data buses of that size supporting 64-bit data types. Keeping abreast with the current trend, we should expect 128-bit or even 256-bit computing to become the household norm eliminating the need for the additional multiple-precision libraries to avail our technique. 

In the fast marching and the fast sweeping methods the incurred error is lower bounded by the grid width as they involve spatial discretization of the derivative operator. Unless we resort to a more finer grid---in the process increasing the number of grid locations $N$---the correctness of these methods cannot be ameliorated. Our method escapes this difficulty altogether as it doesn't require discretization. For a given grid consisting of $N$ grid locations, the accuracy only depends on the value of $\tau$ allowed by the underlying computing architecture which---as we argued---can be progressively set to smaller value with furtherance in technology. 

\section{Fast computation of signed distance functions}
\label{sec:SignedDist}
The solution for the approximate Euclidean distance function in (\ref{STilde})
is lacking in one respect: there is no information regarding the sign of
the distance function. This is to be expected since the distance function was
obtained only from a set of \emph{points} $Y$ and not a closed curve (in 2D) or
surface (in 3D). We now describe a new method for computing the signed distance
function in 2D using winding numbers and in 3D using the topological
degree.

\subsection{Computing winding numbers}

Assume that we have a closed, parametric curve $\left\{ x^{(1)}(t),x^{(2)}(t)\right\} ,\, t\in[0,1]$.
We seek to determine if a grid location in the set $\left\{
X_{i}\in\mathbb{R}^{2},\, i\in\left\{ 1,\ldots,N\right\} \right\} $ 
is inside the closed curve. The winding number is the number of times
the curve winds around the point $X_{i}$ (if at all) with counterclockwise
turns counted as positive and clockwise turns as negative. If a point
is inside the curve, the winding number is a non-zero integer. If
the point is outside the curve, the winding number is zero. If we
can efficiently compute the winding number for all points on a grid
w.r.t. to a closed curve, then we would have the sign information (inside/outside)
for all the points. We now describe a fast algorithm to achieve this
goal.

The change in angle $\theta(t)$ of the curve is given by
$d\theta(t)=\left(\frac{x^{(1)}\dot{x}^{(2)}-x^{(2)}\dot{x}^{(1)}}{\|x\|^{2}}\right)dt$. 
Since we need to determine whether the curve winds around each of
the points $X_{i},i\in\{1,\ldots,N\}$, define
$(\hat{x}_{i}^{(1)},\hat{x}_{i}^{(2)})\equiv(x^{(1)}-X_{i}^{(1)},x^{(2)}-X_{i}^{(2)}),\,\forall
i$. 
Then the winding number for the grid point $X_i$ is  
\begin{equation}
\mu_{i}=\frac{1}{2\pi}\oint_{C}\left(\frac{\hat{x}_{i}^{(1)}\dot{\hat{x}}_{i}^{(2)}-\hat{x}_{i}^{(2)}\dot{\hat{x}}_{i}^{(2)}}{\|\hat{x}_{i}\|^{2}}\right)dt,\,\forall 
i\in\left\{ 1,\ldots,N\right\} .
\label{eq:windingnumbers}
\end{equation}
 In our case, we have a piecewise linear curve
defined by a sequence of points \newline
$\left\{ Y_{k}\in\mathbb{R}^{2},k\in\left\{ 1,\ldots,K\right\} \right\} $. As we also assume that the curve is closed,
the {}``next'' point after $Y_{K}$ is $Y_{1}$. The integral
in (\ref{eq:windingnumbers}) then becomes a discrete summation and we
get 
\begin{equation}
\mu_{i}=\frac{1}{2\pi}\sum_{k=1}^{K}\frac{\left(Y_{k}^{(1)}-X_{i}^{(1)}\right)\left(Y_{k\oplus1}^{(2)}-Y_{k}^{(2)}\right)-\left(Y_{k}^{(2)}-X_{i}^{(2)}\right)
  \left(Y_{k\oplus1}^{(1)}-Y_{k}^{(1)}\right)}{\|Y_{k}-X_{i}\|^{2}} 
\label{eq:kappadiscrete}
\end{equation}
$\forall i\in\left\{ 1,\ldots,N\right\} $, where the notation $Y_{k\oplus1}^{(\cdot)}$
denotes that $Y_{k\oplus1}^{(\cdot)}=Y_{k+1}^{(\cdot)}$ for $k\in\{1,\ldots,K-1\}$
and $Y_{K\oplus1}^{(\cdot)}=Y_{1}^{(\cdot)}$. We can simplify the
notation in (\ref{eq:kappadiscrete}) (and obtain a measure of conceptual
clarity as well) by defining the {}``tangent'' vector $\left\{ Z_{k},k=\left\{ 1,\ldots,K\right\} \right\} $
as 
\begin{equation}
Z_{k}^{(\cdot)}=Y_{k\oplus1}^{(\cdot)}-Y_{k}^{(\cdot)},k\in\left\{
1,\ldots,K\right\} 
\label{eqn:tangentVec}
\end{equation}
with the $(\cdot)$ symbol indicating either coordinate. Using the
tangent vector $Z_k$, we rewrite (\ref{eq:kappadiscrete}) as 
\begin{equation}
\mu_{i}=\frac{1}{2\pi}\sum_{k=1}^{K}\frac{\left(Y_{k}^{(1)}-X_{i}^{(1)}\right)Z_{k}^{(2)}-\left(Y_{k}^{(2)}-X_{i}^{(2)}\right)Z_{k}^{(1)}}{\|Y_{k}-X_{i}\|^{2}},\,\forall
i\in\left\{ 1,\ldots,N\right\} 
\label{eq:kappdiscreteZ}
\end{equation}
We observe that $\mu_i$ in (\ref{eq:kappdiscreteZ})
is a sum of two discrete convolutions. The first convolution is between
two functions $f_{cr}(X)\equiv f_{c}(X)f_{r}(X)$ and $g_{2}(X)=\sum_{k=1}^{K}Z_{k}^{(2)}\delta_{\mathrm{kron}}(X-Y_{k})$.
The second convolution is between two functions $f_{sr}(X)\equiv f_{s}(X)f_{r}(X)$ and $g_{1}(X)\equiv\sum_{k=1}^{K}Z_{k}^{(1)}\delta_{\mathrm{kron}}(X-Y_{k})$.
The Kronecker delta function $\delta_{\mathrm{kron}}(X-Y_k)$ is defined (\ref{eq:krondelta}).
The functions $f_{c}(X),\, f_{s}(X)$ and $f_{r}(X)$ are defined
as 
\begin{equation}
f_{c}(X) \equiv \frac{X^{(1)}}{\|X\|},\, f_{s}(X)\equiv\frac{X^{(2)}}{\|X\|}, \mbox{  and   } f_{r}(X) \equiv \frac{1}{\|X\|}.
\label{eq:fcfsfr}
\end{equation}
 Here we have abused the notation and let $X^{(1)}$ ($X^{(2)}$) denote
the $x$ ($y$)-coordinate of the grid point $X$.
Armed with these relationships, we rewrite (\ref{eq:kappdiscreteZ})
to get 
\begin{equation}
\mu(X)=\frac{1}{2\pi}\left[-f_{cr}(X)\ast g_{2}(X)+f_{sr}(X)\ast g_{1}(X)\right]
\label{eq:kappadiscretefinal}
\end{equation}
which can be computed in $O(N\log N)$ time using FFT-based convolution
\emph{simultaneously} for all the $N$ grid points $\left\{X_{i},i=\left\{ 1,\ldots,N\right\} \right\}$.

It is apparent from the definitions given in (\ref{eq:fcfsfr}) that the
functions $f_c(X)$, $f_s(X)$ and $f_r(X)$ are not well-defined at the origin
$X=0$. The convolution of $f_{cr}(X)$ and $f_{sr}(X)$ with the functions
$g_1(X)$ and $g_2(X)$ in (\ref{eq:kappadiscretefinal}) will lead to
ill-defined values for $\mu(X)$ at the source locations $\{Y_k\}_{k=1}^K$. The
$\mu(X)$ values remain unscathed at other grid locations as the convolution
operation just centers the $f_{cr}(X)$ and $f_{sr}(X)$ functions around the
source points.  But, as the winding numbers are not even defined at the source
locations (they are defined only with respect to them), the seemingly ominous
aforementioned problem does not pose any serious threat. In actuality, the
functions $f_c(X)$, $f_s(X)$ and $f_r(X)$ can take on any arbitrary value at
$X=0$ leading to arbitrary values for $\mu(X)$ at the source locations, which
is meaningless and needs to be ignored. Hence without loss of generality we
set $f_{c}(0)=f_{s}(0)=f_{r}(0)=0$.  

\subsection{Computing topological degree}

The winding number concept for 2D admits a straight forward generalization
to 3D and higher dimensions. The equivalent concept is the topological
degree which is based on normalized flux computations. Assume that
we have an oriented surface in 3D \cite{Gray97} which is represented
as a set of $K$ triangles. The $k^{th}$ triangle has an outward
pointing normal $P_{k}$ and this can easily be obtained once the
surface is oriented. (We vectorize the edge of each triangle. Since
triangles share edges, if the surface can be oriented, then there's
a consistent way of lending direction to each triangle edge. The triangle
normal is merely the cross-product of the triangle vector edges.)
We pick a convenient triangle center (the triangle incenter for instance)
for each triangle and call it $Y_{k}$. The normalized flux (which
is very closely related to the topological degree) \cite{Aberth98}
determines the ratio of the outward flux from a point $X_{i}$ treated
as the origin. If $X_{i}$ is outside the enclosed surface, then the
total outward flux is zero. If the point is inside, the outward normalized
flux will be non-zero and positive.

The normalized flux for a point $X_{i}$ is 
\begin{equation}
\mu_{i}=\frac{1}{4\pi}\sum_{k=1}^{K}\frac{\langle(Y_{k}-X_{i}),P_{k}\rangle}{\|Y_{k}-X_{i}\|^{3}}.
\label{eq:normflux}
\end{equation}
 This can be written in the form of convolutions. To see this we
write (\ref{eq:normflux}) in component form: 
\begin{equation}
\mu_{i}=\frac{1}{4\pi}\sum_{k=1}^{K}\frac{(Y_{k}^{(1)}-X_{i}^{(1)})P_{k}^{(1)}+(Y_{k}^{(2)}-X_{i}^{(2)})P_{k}^{(2)}+(Y_{k}^{(3)}-X_{i}^{(3)})P_{k}^{(3)}}{\|Y_{k}-X_{i}\|^{3}}
\label{eq:mui}
\end{equation}
 which can be simplified as 
 \begin{equation}
\mu(X)=-\frac{1}{4\pi}\left(f_{1}(X)\ast g_{1}(X)+f_{2}(X)\ast g_{2}(X)+f_{3}(X)\ast g_{3}(X)\right)
\label{eq:mudef}
\end{equation}
where $f_{(\cdot)}(X)\equiv\frac{X^{(\cdot)}}{\|X\|^{3}}$ and
$g_{(\cdot)}(X)\equiv\sum_{k=1}^{K}P_{k}^{(\cdot)}\delta_{\mathrm{kron}}(X-Y_{k})$. The
Kronecker delta function $\delta_{\mathrm{kron}}(X-Y_k)$ is defined (\ref{eq:krondelta}). This can be computed 
in $O(N\log N)$ time using FFT-based convolution for all the $N$
grid points $X_{i}$ \emph{simultaneously}.

For the sake of clarity we explicitly show the generalization of
the winding number to the topological degree by rewriting some of
the calculations involved in computing the winding number. Recall
that for every point $Y_{k}$ on the discretized curve, we defined
its tangent vector $Z_{k}$ in (\ref{eqn:tangentVec}).
The \emph{outward pointing normal} $P_{k}=(P_{k}^{(1)},P_{k}^{(2)})$,
at the point $Y_{k}$ ($P_{k}$ will point outwards provided $Y_{1},Y_{2},\ldots,Y_{k}$
are taken in the anti-clockwise order), is given by $P_{k}^{(1)}=Z_{k}^{(2)},P_{k}^{(2)}=-Z_{k}^{(1)}$.
Using the normal vector $P_{k}$, (\ref{eq:kappdiscreteZ})
can be rewritten as 
\begin{equation}
\mu_{i}=\frac{1}{2\pi}\sum_{k=1}^{K}\frac{\langle(Y_{k}-X_{i}),P_{k}\rangle}{\|Y_{k}-X_{i}\|^{2}}.\label{eq:normflux_2D}
\end{equation}
 The similarity between (\ref{eq:normflux_2D}) and (\ref{eq:normflux})
 is conspicuous. This lends extra validity to the fact that the topological degree is just a generalization of the winding number concept. It is important to bear in mind that the computation of the winding numbers and the topological degree \emph{does not} require arbitrary precision package as no exponentially decaying functions are involved.

\section{Fast computation of the derivatives of the distance function\label{sec:Fastderivatives}}

Just as the approximate Euclidean distance function $S(X)$
can be efficiently computed in $O(N\log N)$, so can the derivatives.
This is important because fast computation of the derivatives of $S(X)$
on a regular grid can be very useful in curvature
computations. Below, we detail how this can be achieved. We begin
with the gradients and for illustration purposes, the derivations
are performed in 2D: 
\begin{equation}
S_{x}(X)=\frac{\sum_{k=1}^{K}\frac{X^{(1)}-Y_{k}^{(1)}}{\|X-Y_{k}\|}\exp\left(-\frac{\|X-Y_{k}\|}{\tau}\right) }{\sum_{k=1}^{K}\exp\left( -\frac{\|X-Y_{k}\|}{\tau}\right) }.
\label{eq:SxSy}
\end{equation}
 A similar expression can be obtained for $S_{y}(X)$. These
first derivatives can be rewritten as discrete convolutions: 
\begin{equation}
S_{x}(X)=\frac{f_{c}(X)f(X)\ast g(X)}{f(X)\ast g(X)},\, S_{y}(X)=\frac{f_{s}(X)f(X)\ast g(X)}{f(X)\ast g(X)},
\label{eq:SxSyconv}
\end{equation}
 where $f_{c}(X)$ and $f_{s}(X)$ are as defined in (\ref{eq:fcfsfr})
and $f(X)$ and $g(X)$ are given in (\ref{eq:function_f}) and (\ref{eq:function_g}) respectively.

The second derivative formulas are somewhat involved. Rather than
hammer out the algebra in a turgid manner, we merely present the final
expressions---all discrete convolutions---for the three second derivatives
in 2D:
\begin{eqnarray}
S_{xx}(X) & = & \frac{\left[-\frac{1}{\tau}f_{c}^{2}(X)+f_{s}^{2}(X)f_{r}(X)\right]f(X)\ast g(X)}{f(X)\ast g(X)}+\frac{1}{\tau}(S_{x})^{2}(X),\label{eq:Sxx}\\
S_{yy}(X) & = &
\frac{\left[-\frac{1}{\tau}f_{s}^{2}(X)+f_{c}^{2}(X)f_{r}(X)\right]f(X)\ast
  g(X)}{f(X)\ast g(X)}+\frac{1}{\tau}(S_{y})^{2}(X),\mathrm{\,
  and}\nonumber\\ & &\label{eq:Syy}\\
S_{xy}(X) & = &
\frac{-\left[\frac{1}{\tau}+f_{r}(X)\right]f_{c}(X)f_{s}(X)f(X)\ast
  g(X)}{f(X)\ast
  g(X)}+\frac{1}{\tau}S_{x}(X)S_{y}(X)\nonumber\\ & &\label{eq:Sxy}
\end{eqnarray}
 where $f_{r}(X)$ is as defined in (\ref{eq:fcfsfr}). Since we can efficiently compute the first and second derivatives
of the approximate Euclidean distance function everywhere on a regular
grid, we can also compute derived quantities such as curvature (Gaussian,
mean and principal curvatures) for the two-dimensional surface $S(X)$
computed at the grid locations $X$. In Section~\ref{sec:Experiments} we visualize
the derivatives for certain shape silhouettes.

\subsection{Convergence analysis for the derivatives}
\label{sec:derivConvergence}
We now show convergence of the derivatives ($S_x,S_y$) obtained via
(\ref{eq:SxSy}) to their true value as $\tau \rightarrow 0$. Recall that the
true distance function is not differentiable at the point-source locations
$\{Y_k\}_{k=1}^K$ and on the Voronoi boundaries which corresponds to grid
locations which are equidistant from two or more point sources. Hence it is
meaningful to establish convergence only for grid locations whose closest
source-point can be uniquely determined.  

For the purposes of illustration we show the analysis in 2D. Let $Y_{k_0}$ be the
unique closest source-point for a grid location $X = (X^{(1)},X^{(2)})$, i.e,
$R(X)=\|X-Y_{k_0}\| < \|X-Y_k\|, \forall k \not=k_0$. Then the true
derivatives at the location $X$ are given by 
\begin{equation}
R_x(X) = \frac{X^{(1)}-Y_{k_0}^{(1)}}{R(X)} , R_y(X)  = \frac{X^{(2)}-Y_{k_0}^{(2)}}{R(X)}.
\end{equation} 
Multiplying and dividing the (\ref{eq:SxSy}) by $\exp\left(\frac{R(X)}{\tau}\right)$, we get
\begin{equation}
S_x(X) = \frac{\frac{X^{(1)}-Y_{k_0}^{(1)}}{R(X)}+
  \sum_{k=1,k\not=k_0}^{K}\frac{X^{(1)}-Y_{k}^{(1)}}{\|X-Y_{k}\|}\exp\left(-\frac{\gamma_k}{\tau}\right)}{1
  +\sum_{k=1,k\not=k_0}^{K}\exp\left(-\frac{\gamma_k}{\tau}\right) } 
\label{eq:Sx1}
\end{equation}
where $\gamma_k = \|X-Y_{k}\|-R(X)$. Since $\gamma_k >0, \forall k \not=k_0$,
it follows that $\lim_{\tau \rightarrow 0}
\exp\left(-\frac{\gamma_k}{\tau}\right)=0, \forall k\not=k_0$. Since all the
other terms in (\ref{eq:Sx1}) are independent of $\tau$ we get $\lim_{\tau
  \rightarrow 0} S_x(X)=R_x(X)$. The convergence analysis for $S_y(X)$ to
$R_y(X)$ follows along similar lines.  

Furthermore, the gradient magnitude $\|\nabla S(X)\|$ for any non-zero value
of $\tau$ will be strictly less than its true value
$\sqrt{(R_x(X))^2+(R_y(X))^2} = 1$. To see this consider (\ref{eq:SxSy}) and
for the sake of convenience define $\alpha_k =
\frac{X^{(1)}-Y_{k}^{(1)}}{\|X-Y_{k}\|}$, $\beta_k =
\frac{X^{(2)}-Y_{k}^{(2)}}{\|X-Y_{k}\|}$ and $E_k =
\exp\left(-\frac{\|X-Y_{k}\|}{\tau}\right)$. Since $\alpha_k^2+\beta_k^2 = 1,
\forall k$, after simple algebraic manipulations we get 
\begin{equation}
\|\nabla S(X)\|^2 = \frac{\sum_{k=1}^K E_k^2 + 2 \sum_{k=1}^K \sum_{l>k}^K
  \left(\alpha_k \alpha_l+\beta_k \beta_l\right)E_k E_l}{\sum_{k=1}^K
  E_k^2+2\sum_{k=1}^K \sum_{l>k}^K E_k E_l}. 
\end{equation}
Since $k \not= l$, from the Cauchy-Schwarz inequality we have $\left|\alpha_k \alpha_l+\beta_k \beta_l\right| < 1$.
It is then easy to see that $\|\nabla S(X)\| < 1$. We also provide
experimental evidence in the subsequent section to corroborate this
fact. Nevertheless, the magnitude value converges to 1 as $\tau \rightarrow 0$
at all the grid locations (barring the point-sources and the Voronoi
boundaries) as the gradients themselves converge to their true values.  

\section{Experiments}
\label{sec:Experiments}
In this section we demonstrate the usefulness of our fast convolution approach to computing Euclidean distance functions
on a bounded 2D and 3D grid. As we discussed before in
Section~\ref{sec:numericalissues} we go beyond the precision
supported by the double floating-point numbers (64 bits) and resort to
the GNU multiple-precision library (GMP) and MPFR arbitrary precision packages to
improve the computational accuracy of our technique. 
For the following experiments we used $p=512$ precision bits.

\subsection{2D Experiments}
\begin{center}
\begin{figure}
\begin{minipage}[c]{0.45\linewidth}	
\centering
\includegraphics[clip,width=1\textwidth]{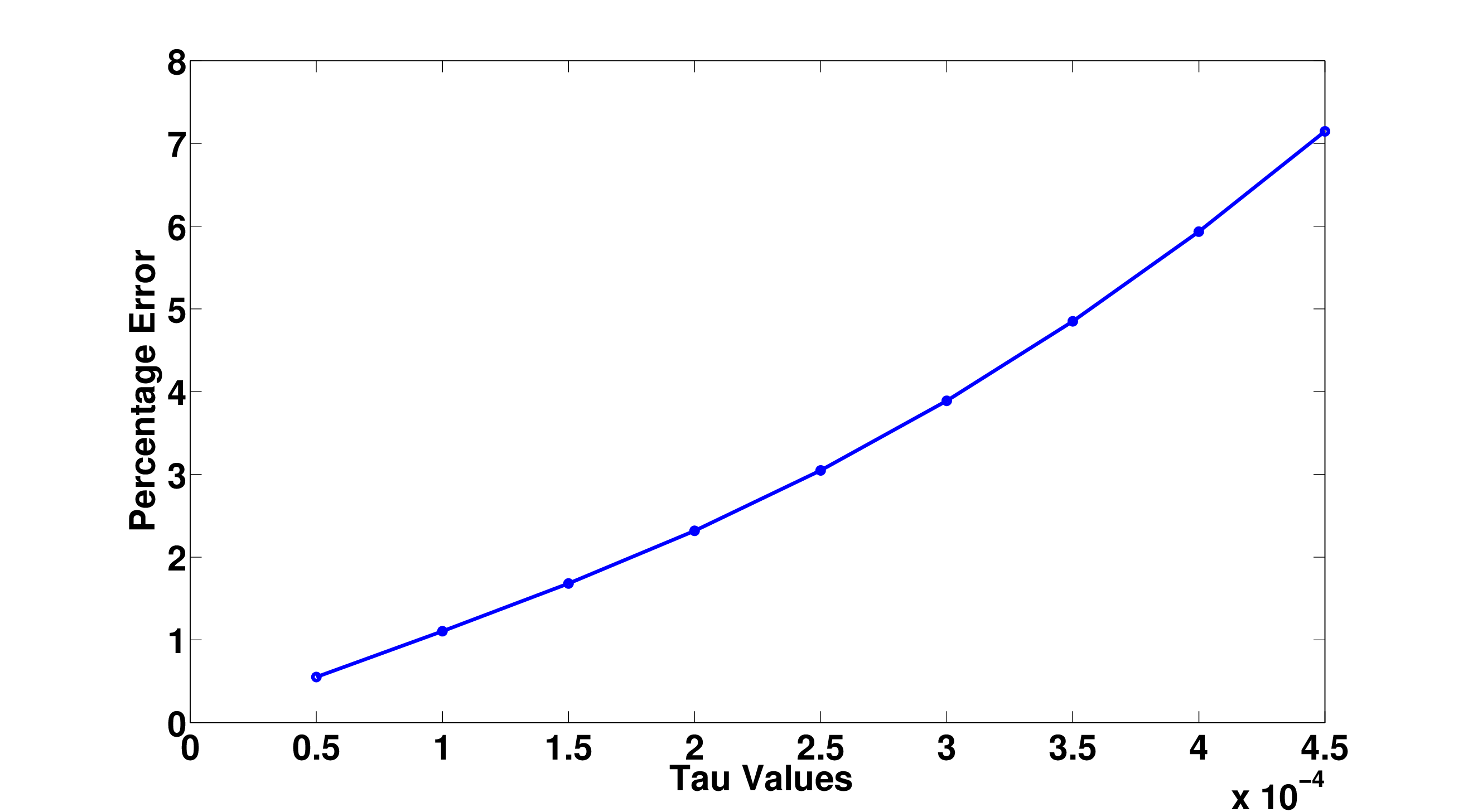} 
\caption{Mean percentage error versus $\tau$ in 1000 2$D$ experiments.}
\label{fig:difftauVals}
\end{minipage}%
\begin{minipage}[c]{0.45\linewidth}	
\centering
\includegraphics[clip,width=1\textwidth]{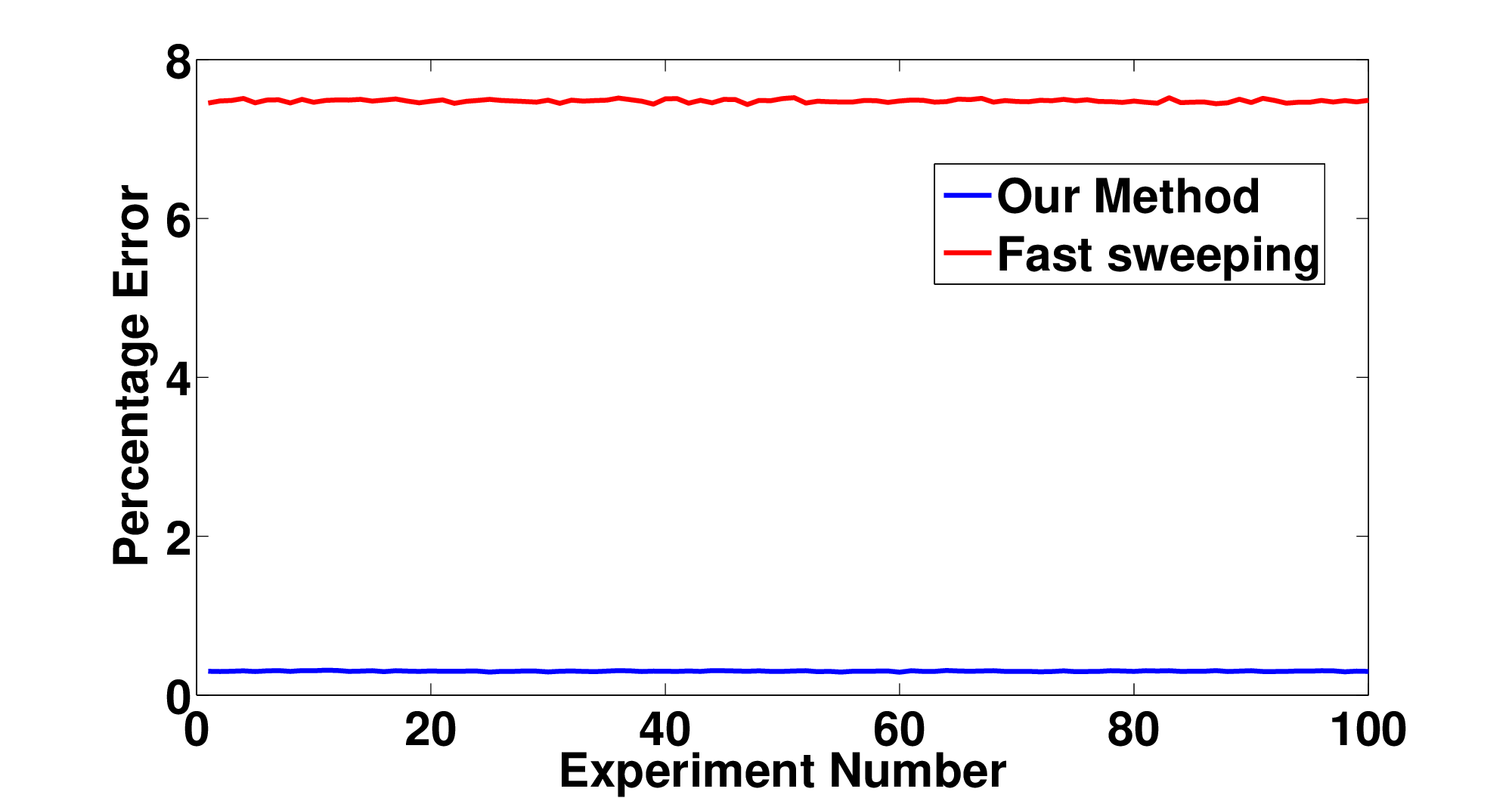} 
\caption{Percentage error between the true and computed Euclidean distance
function for (i) Our method (in blue) (ii) Fast sweeping (in red) in 100 $2D$ experiments. 
\label{fig:SchFSRandomLoc}}
\end{minipage}%
\end{figure}
\end{center}

\vspace{0.1cm}
\noindent \textbf{Example 1:} We begin by discussing the effect
of $\tau$ on our method and establish that as $\tau\rightarrow0$
the accuracy our method empirically improves. To this end, we
considered a 2D grid consisting of points between $-0.125\leq x\leq 0.125$ and $-0.125\leq y\leq 0.125$ at a grid width of $\frac{1}{2^{9}}$.
We ran $1000$ experiments randomly choosing $5000$ grid locations
as data points (point-set), for $9$ different values of $\tau$
ranging from $5\times10^{-5}$ to $4.5\times10^{-4}$ in steps of
$5\times10^{-5}$. For each execution and at each value of $\tau$ we calculated
the percentage error as 
\begin{equation}
\mathcal{E}=\frac{100}{N}\sum_{i=1}^{N}\frac{\Delta_{i}}{R_{i}}
\label{eq:percentError}
\end{equation}
 where $R_{i}$ and $\Delta_{i}$ are respectively the actual distance
and the absolute difference of the computed distance to the actual
distance at the $i^{th}$ grid point. Figure~\ref{fig:difftauVals}
shows the \emph{mean} percentage error at each value of $\tau$. Note that the error varies almost linearly with $\tau$ 
in accordance with the maximum error bound of $\tau \log K$ proved in Section~\ref{sec:ErrorBound}.
The \emph{maximum} value of the error at each value of $\tau$
is summarized in Table~\ref{table:maxError}.
The error is less than $0.6$\% at $\tau=0.00005$ demonstrating
the algorithm's ability to compute accurate Euclidean distance functions.

\begin{table}
\begin{minipage}[c]{0.45\linewidth}	
\caption{Maximum percentage error for different values of $\tau$}
 \label{table:maxError}
 \centering
\begin{tabular}{|c|c|}
\hline 
$\tau$  & Maximum error \tabularnewline
\hline 
0.00005 & 0.5728\% \tabularnewline
\hline 
0.0001  & 1.1482\% \tabularnewline
\hline 
0.00015 & 1.7461\% \tabularnewline
\hline
0.0002  & 2.4046\% \tabularnewline
\hline
0.00025 & 3.1550\% \tabularnewline
\hline
0.0003  & 4.0146\% \tabularnewline
\hline
0.00035 & 4.9959\% \tabularnewline
\hline
0.0004  & 6.1033\% \tabularnewline
\hline
0.00045 & 7.3380\% \tabularnewline
\hline
\end{tabular}
\end{minipage}%
\begin{minipage}[c]{0.45\linewidth}	
\caption{Percentage error for the Euclidean distance function computed using
the grid points of these silhouettes as data points}
 \label{table:silhouettesData}
\centering
\begin{tabular}{|l|l|l|}
\hline 
Shape  & Fast convolution  & Fast sweeping \tabularnewline
\hline 
Horse  & 2.659\%  & 2.668\% \tabularnewline
\hline 
Hand  & 2.182\%  & 2.572\% \tabularnewline
\hline 
Bird  & 2.241\%  & 1.895\% \tabularnewline
\hline
\end{tabular}
\end{minipage}%
\end{table}
\vspace{-0.5cm}
\noindent \textbf{Example 2:} We pitted our algorithm against
the fast sweeping method \cite{Zhao05} on a $2D$ grid consisting
of points between $-0.125\leq x\leq 0.125$ and $-0.125\leq y\leq 0.125$ at a
grid spacing of $\frac{1}{2^{10}}$. We ran $100$ experiments, each time randomly
choosing $10,000$ grid points as data points. We set $\tau=0.0001$
for our approach and ran the fast sweeping for
$10$ iterations more than enough for it to converge. The plot in Figure~\ref{fig:SchFSRandomLoc}
shows the average percentage error calculated according to (\ref{eq:percentError})
for both these techniques in comparison to the true Euclidean distance
function. From the plot it is clear that while the fast sweeping
method has a percentage error of around \emph{$7\%$}, our fast convolution method
method gave a percentage error of less than $\mathbf{1.5\%}$ providing
much better accuracy.\\\\
\noindent \textbf{Example 3:} We then executed our algorithm
on a set of 2D shape silhouettes\footnote{We thank Kaleem Siddiqi for
  providing us with the set of 2D shape silhouettes
used in this paper. 
}. We sampled along the silhouette of these 2D shapes to generate the source points and then rounded the point cloud to the nearest grid points.
The grid size is $-0.125\leq x\leq 0.125$ and $-0.125\leq y\leq 0.125$ at
a grid width of $\frac{1}{2^{10}}$. We set $\tau$ for our 
method at $0.0003$. For the sake of comparison we
ran the fast sweeping method for $10$ iterations which was sufficient for it to converge.
The percentage errors calculated according to (\ref{eq:percentError})
for both fast convolution and fast sweeping in comparison to the true Euclidean distance
function for these shapes are summarized
in Table~\ref{table:silhouettesData}. The true Euclidean distance function contour plots and those obtained
from fast convolution and fast sweeping are shown in Figure~\ref{fig:silhouettesContourPlots}.
\begin{center}
\begin{figure}[ht]
\begin{centering}
\includegraphics[width=0.3\textwidth]{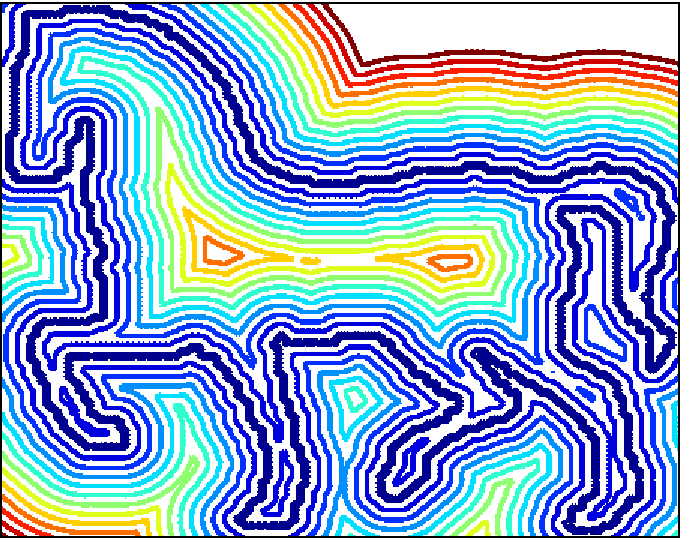}
\includegraphics[width=0.3\textwidth]{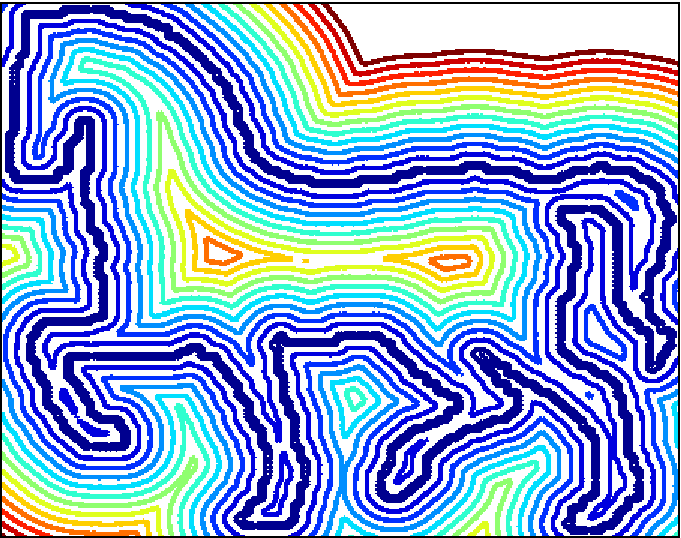}
\includegraphics[width=0.3\textwidth]{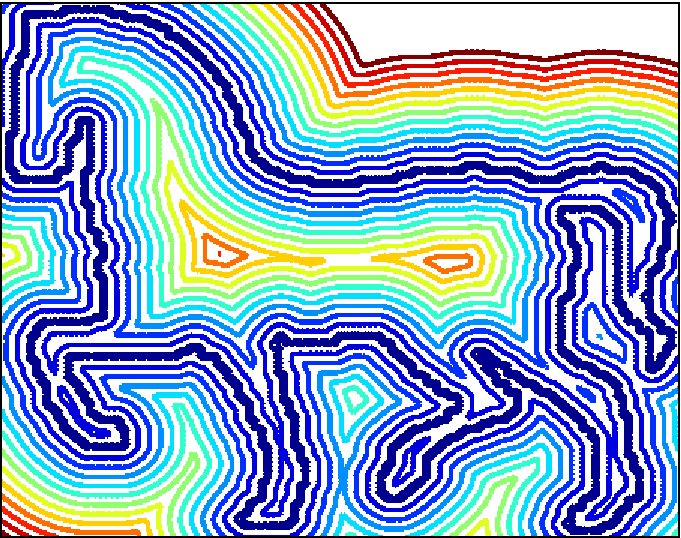}
\includegraphics[width=0.3\textwidth]{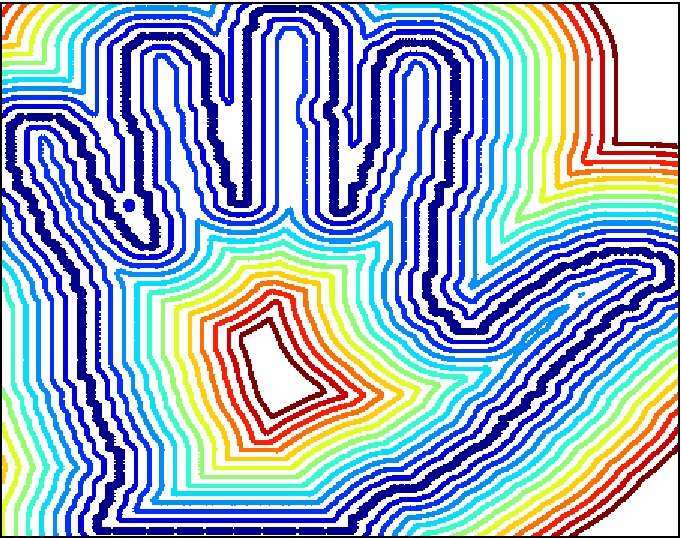}
\includegraphics[width=0.3\textwidth]{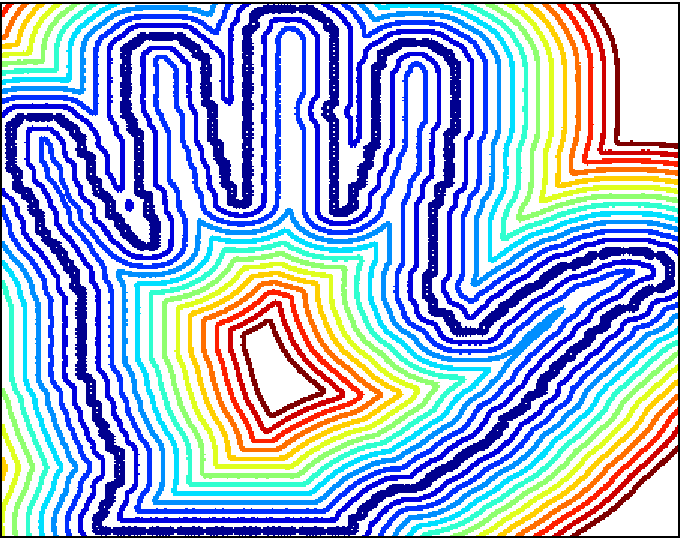}
\includegraphics[width=0.3\textwidth]{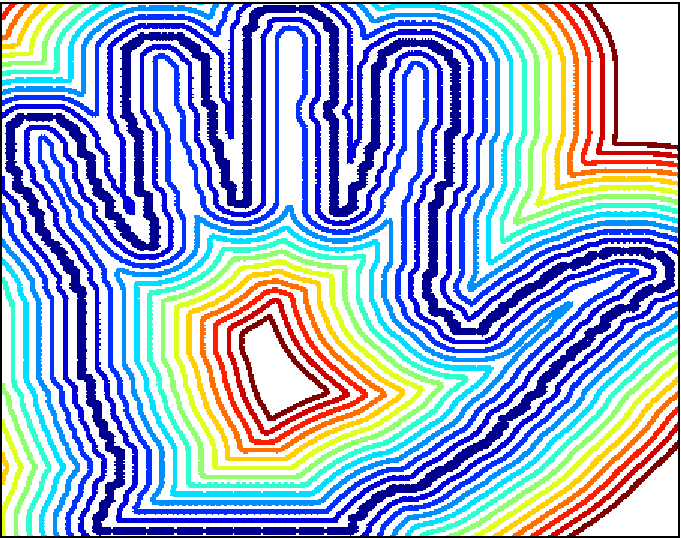}
\includegraphics[width=0.3\textwidth]{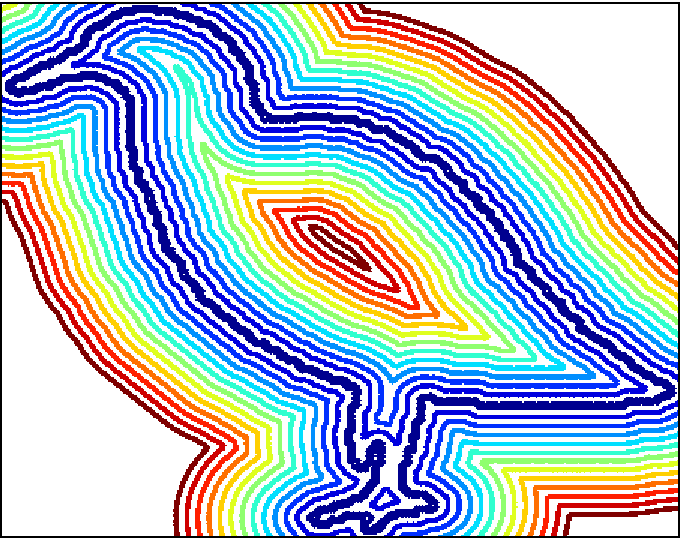}
\includegraphics[width=0.3\textwidth]{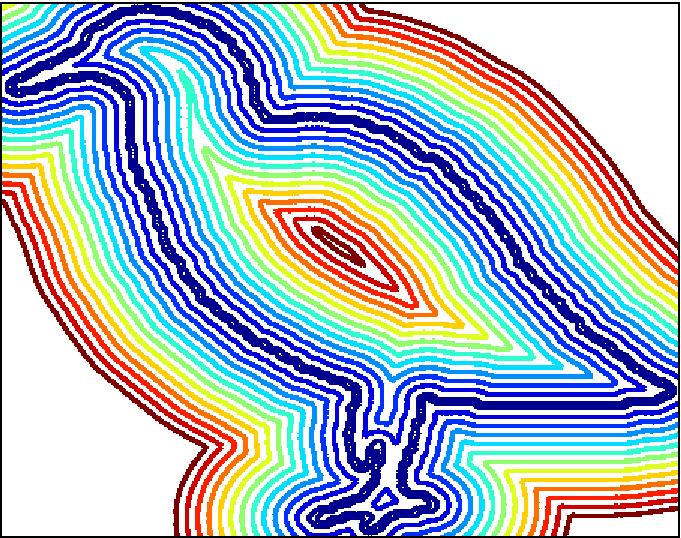}
\includegraphics[width=0.3\textwidth]{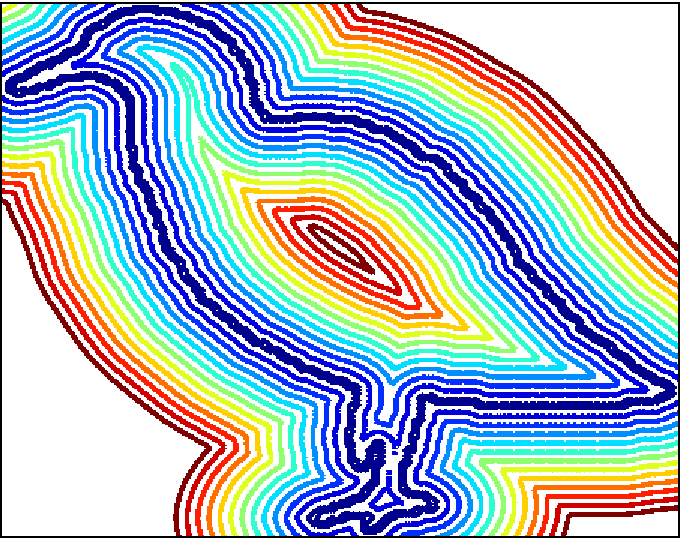} 
\par\end{centering}
\caption{Contour plots: (i) Left: True Euclidean distance function, (ii)
  Center: Fast convolution, (iii) Right: Fast sweeping }
\label{fig:silhouettesContourPlots} 
\end{figure}
\par\end{center}
\vspace{-0.5cm}

In order to differentiate between 
the grid locations that lie inside or outside to these shape silhouettes we
computed the winding number for all the grid points \emph{simultaneously} 
in $O(N \log N)$ using our convolution-based winding number method. 
Grid points with a winding number value greater than 0 after rounding
were marked as interior points. In the the top part of Figure~\ref{fig:Quiverplots}
we visualize the vector fields $(S_{x},S_{y})$
for all the interior points marked in blue (please zoom in to view the gradient vector fields). We see that 
our convolution-based technique for computing the winding number 
separates the interior grid points from the exterior with almost zero
error. In the bottom part of Figure~\ref{fig:Quiverplots} 
we plot the distribution of the winding number values computed over all the
interior and the exterior locations. Observe that for almost all the grid
points, the winding number values are close to \emph{binary}, i.e either 0 or
1, with a value of 0 marking the exterior points (as the curve doesn't wind
around them) and 1 representing interior locations.  
\begin{center}
\begin{figure}
\begin{centering}
\includegraphics[width=0.3\textwidth]{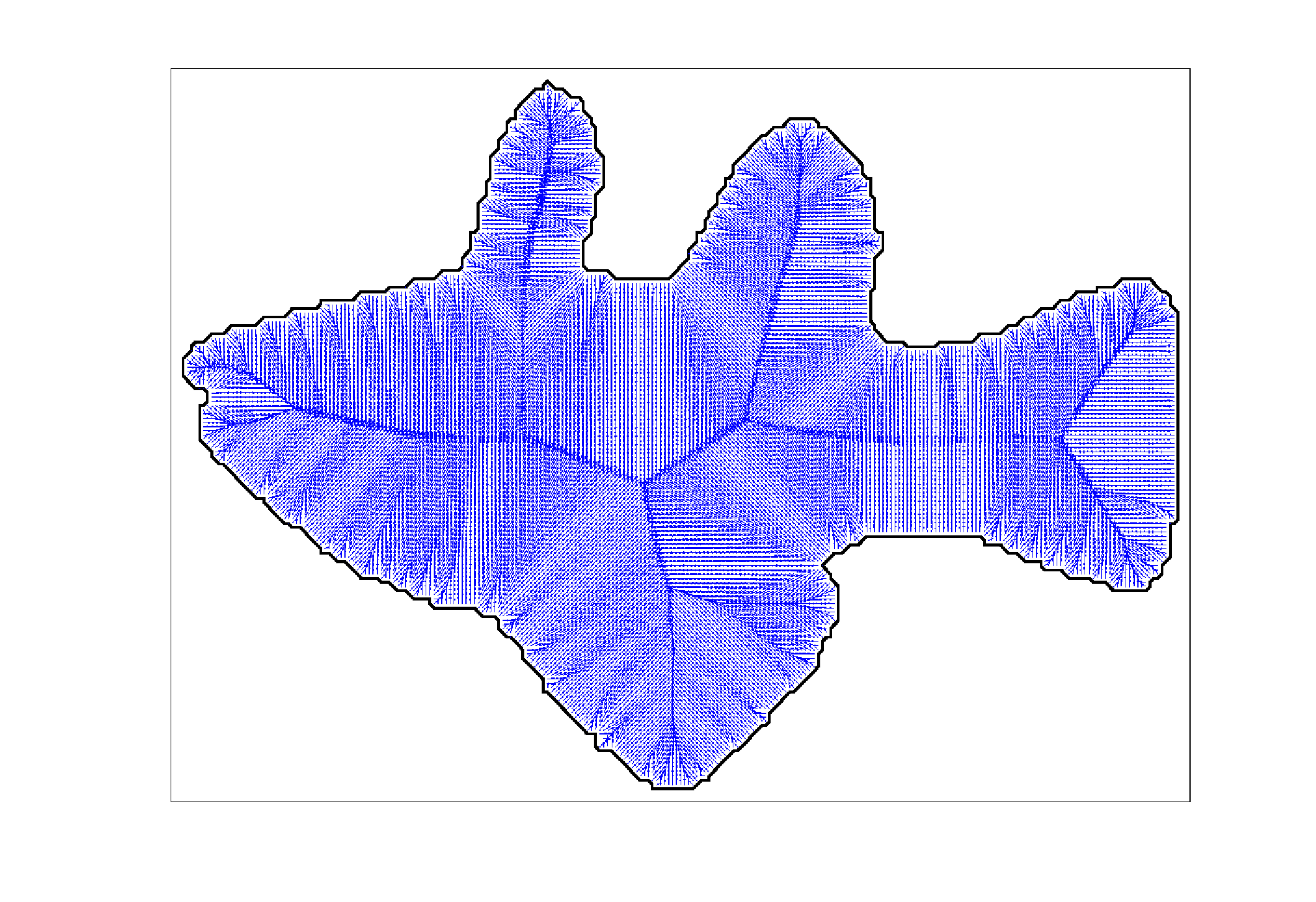}
\includegraphics[width=0.3\textwidth]{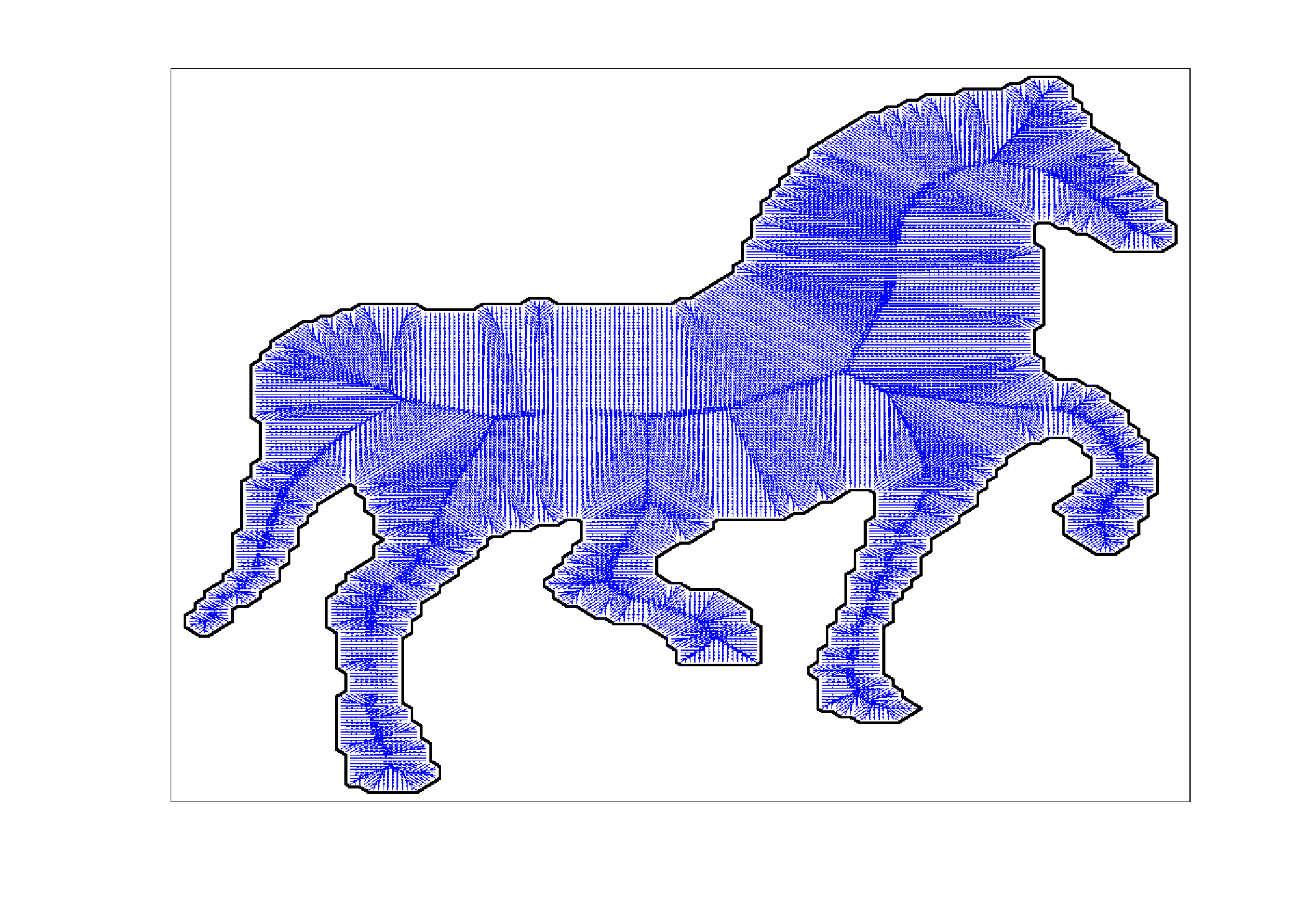}
\includegraphics[width=0.3\textwidth]{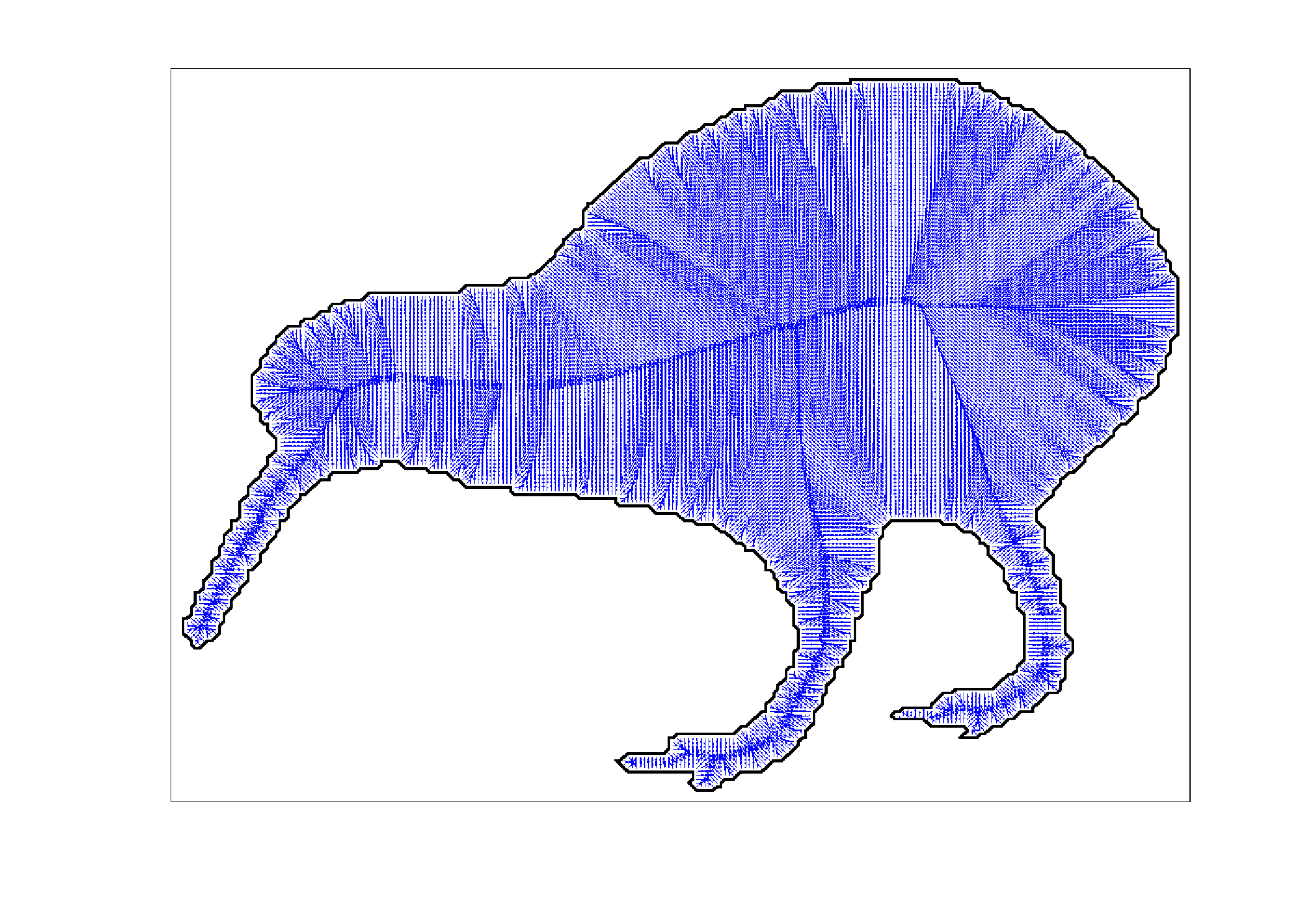}
\includegraphics[width=0.31\textwidth]{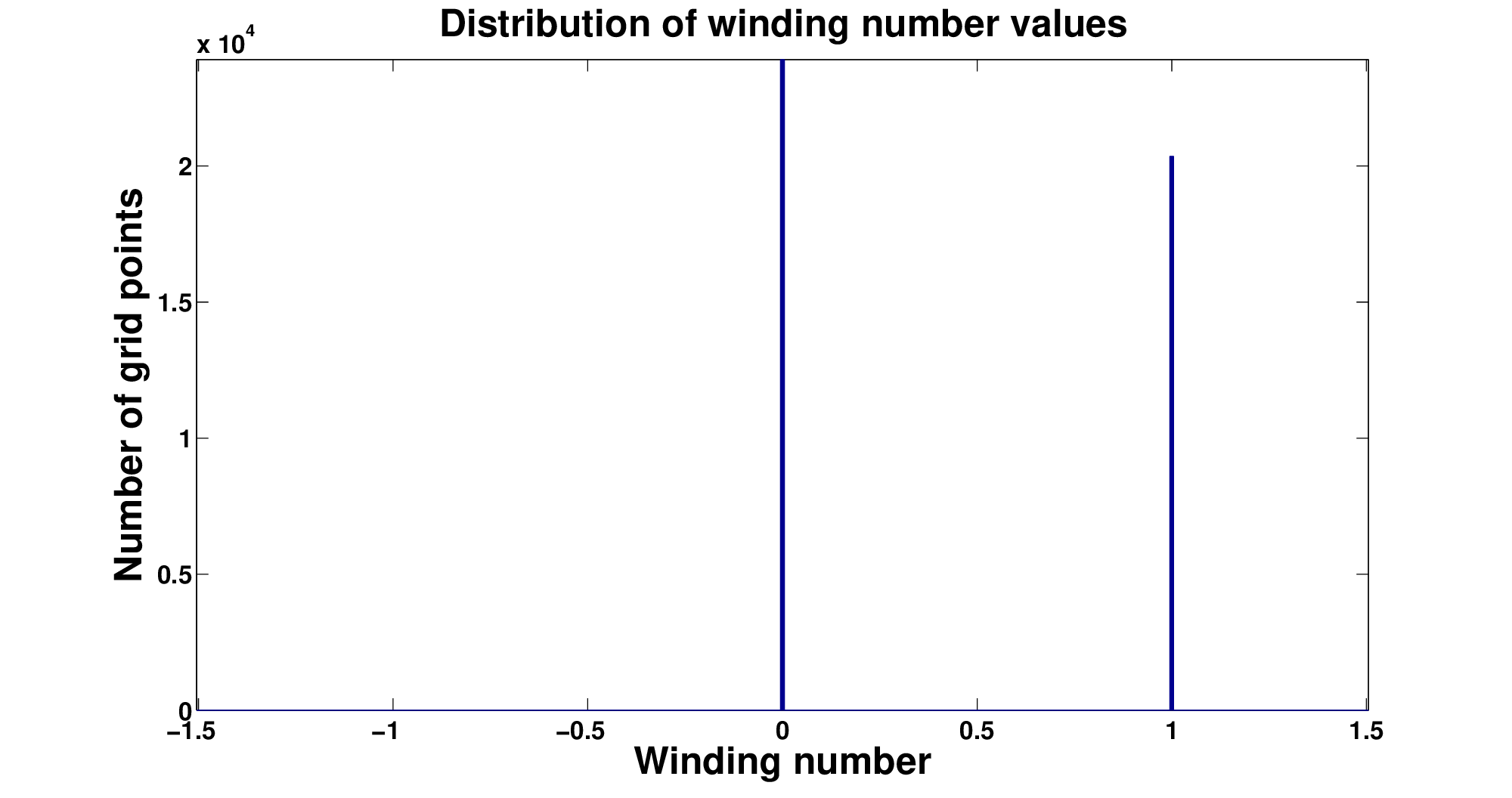}
\includegraphics[width=0.31\textwidth]{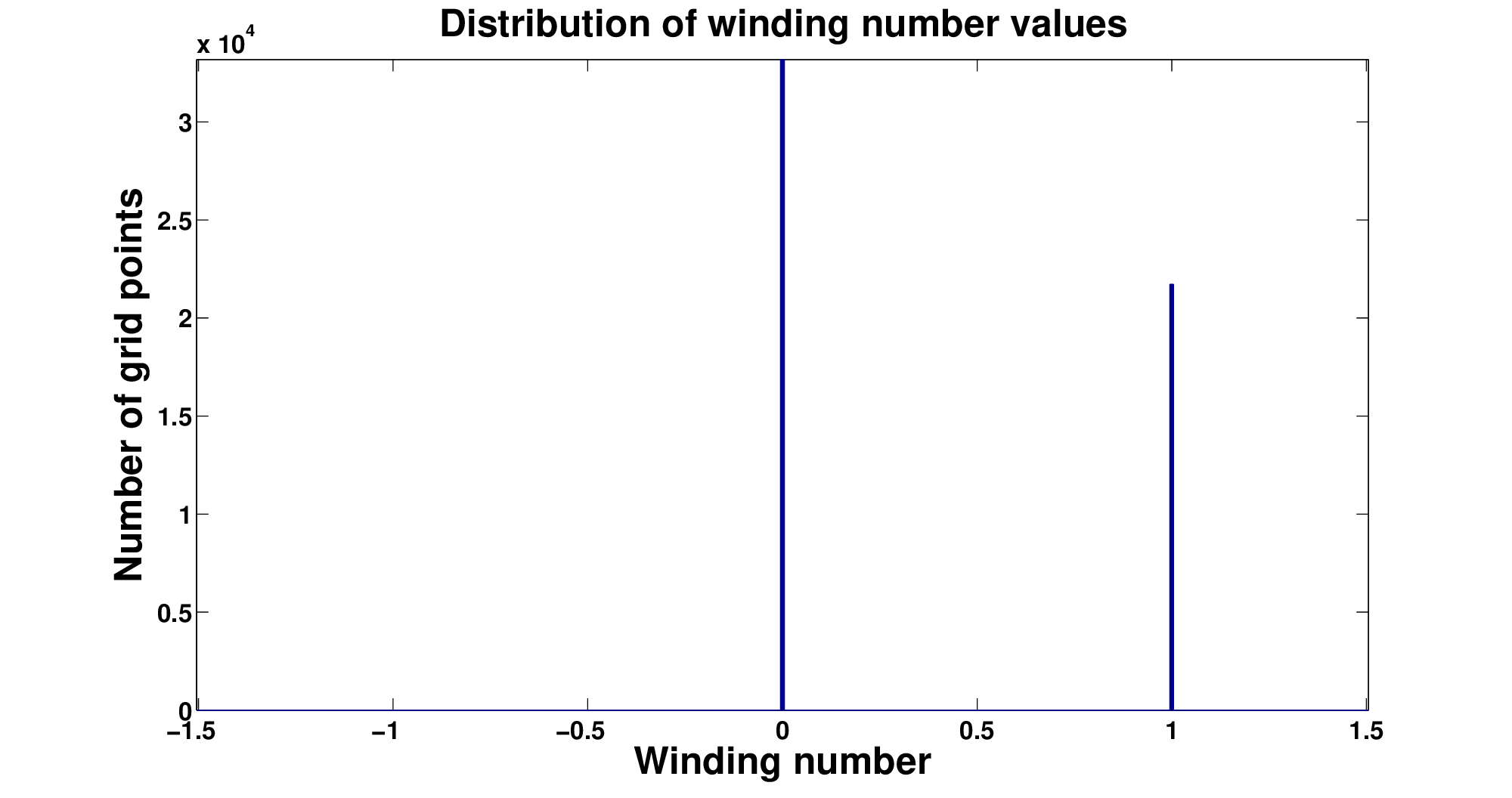}
\includegraphics[width=0.31\textwidth]{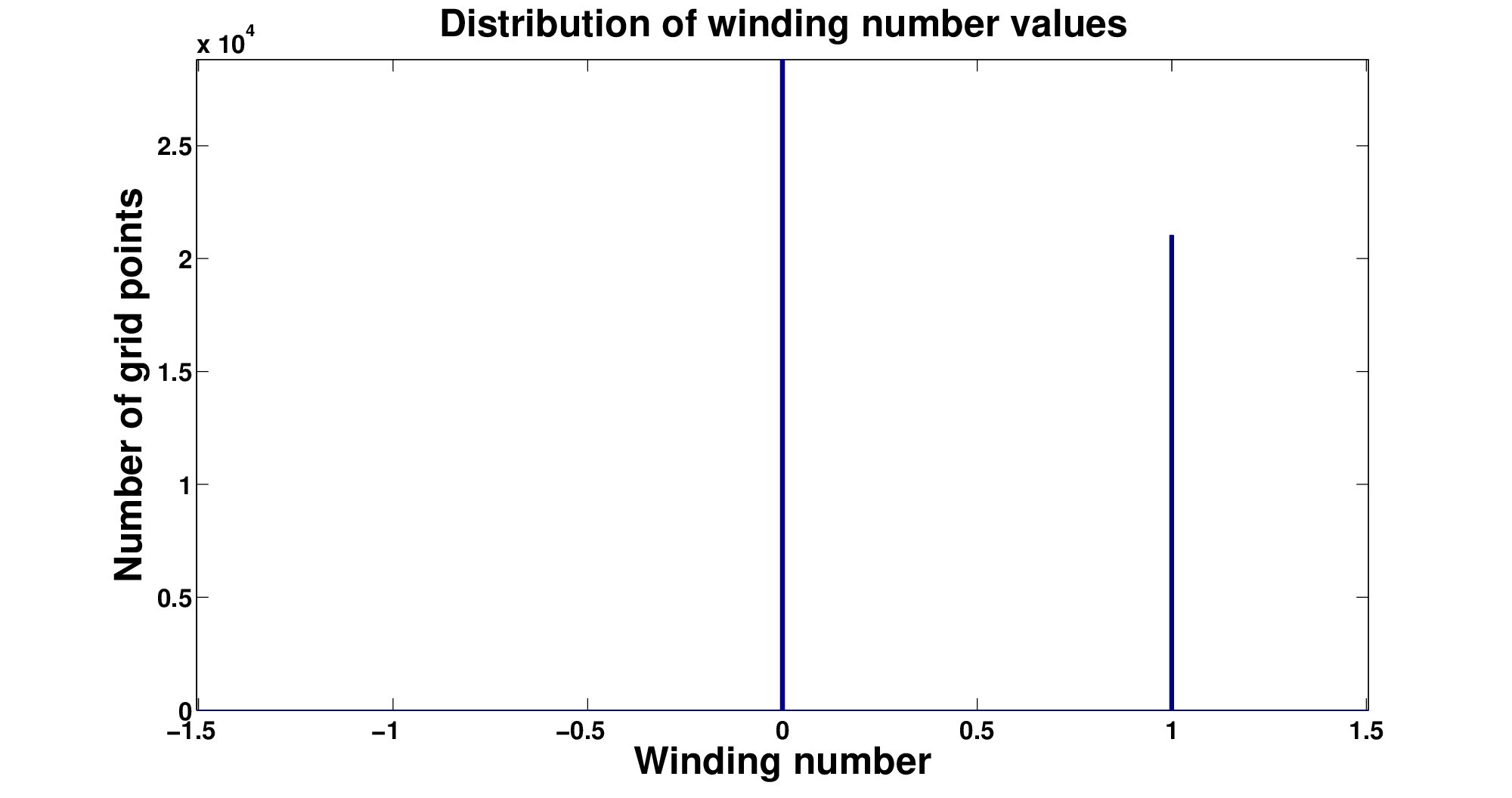}
\par\end{centering}
\caption{(i) Top: Quiver plot of $\nabla S=(S_{x},S_{y})$ for a
set of silhouette shapes (best viewed in color), (ii) Bottom: Distribution of winding numbers
\label{fig:Quiverplots}}
\end{figure}
\par\end{center}
\vspace{-0.9cm}

In the top part of Figure~\ref{fig:Gradmagnitude} we plot the distribution of
the gradient magnitude ($\|\nabla S\|$) values. Since we don't solve for the
true distance function $R$ and rather approximate it by (\ref{STilde}) the
magnitude of the gradients ($S_x,S_y$) obtained via (\ref{eq:SxSy}) may not
necessarily satisfy the Hamilton-Jacobi equation (\ref{EucEikonal}) for a
non-zero value of $\tau$ and hence are not identically equal to 1. The
reformulation of the derivatives given in (\ref{eq:Sx1}) drives this point
home. 
Nevertheless, we do observe that most of the gradient magnitude values (about
$90 \%$) are greater than 0.9. Also notice that all the gradient magnitudes
are less than 1, as explained in Section~\ref{sec:derivConvergence}.

In the bottom part of Figure~\ref{fig:Gradmagnitude}
we visualize the gradient magnitude values as an image plot in tints of
gray. Grid locations whose gradient magnitude values are in proximity to 1 are
marked as white and the black color represent grid points whose gradient
magnitude values are closer to its minimum value on the grid. From these image
plots we see that the grid points which are either too close to the
point-sources or those that lie along the medial axis (corresponding to the
Voronoi boundaries for these shapes) incur maximum errors in their gradient
magnitude values. As these grid locations are almost equidistant from multiple
point-sources many of them contribute substantially to the summation value
$\exp\left( -\frac{\|X-Y_{k}\|}{\tau}\right)$ instead of just the nearest one
leading to a relatively high induced error. From these visualizations we speculate that these inaccuracies when used prudently may actually aid in medial axis computation. We
are currently investigating whether this can be achieved. 
\begin{center}
\begin{figure}
\begin{centering}
\includegraphics[width=0.3\textwidth]{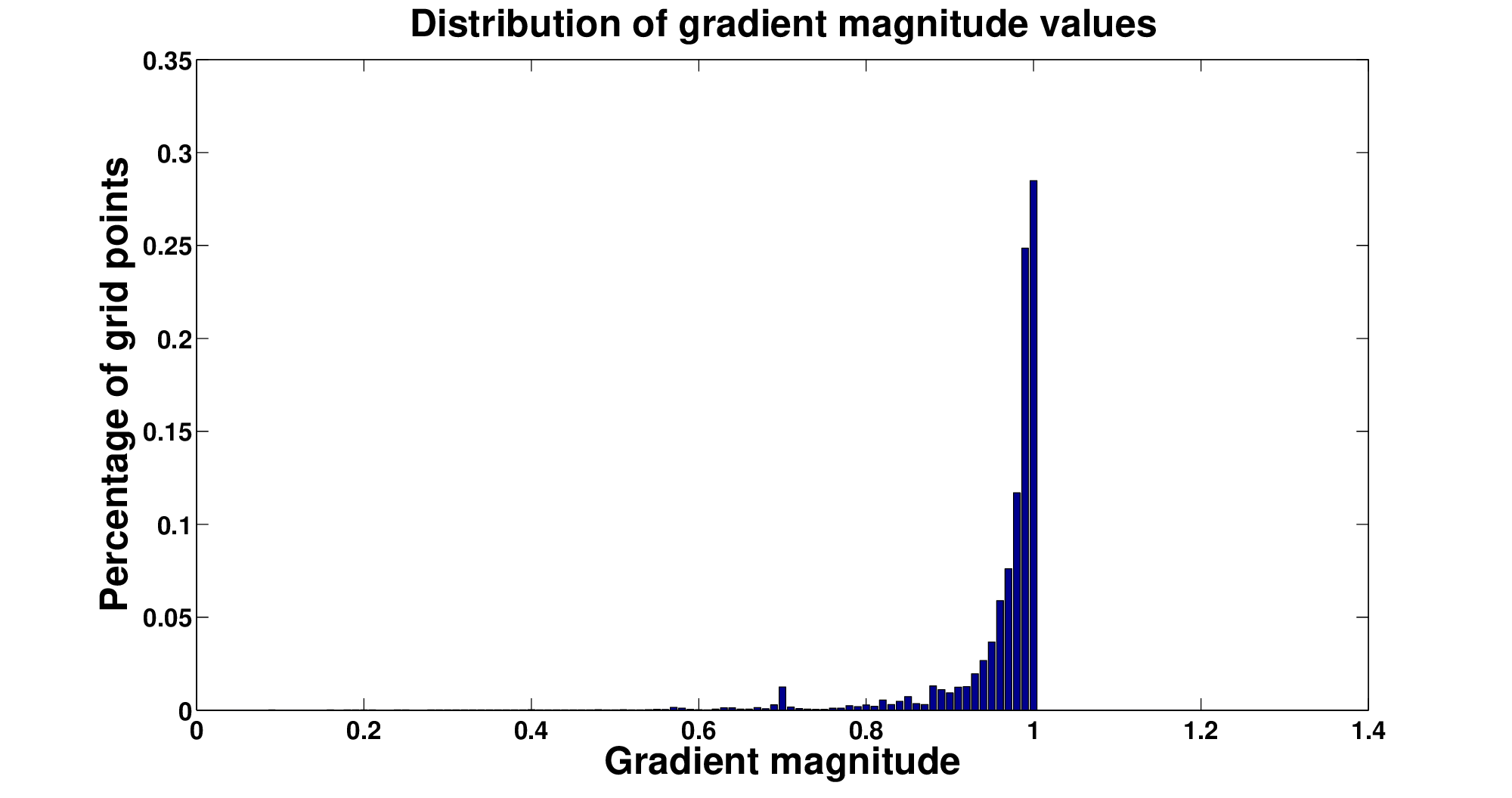}
\includegraphics[width=0.3\textwidth]{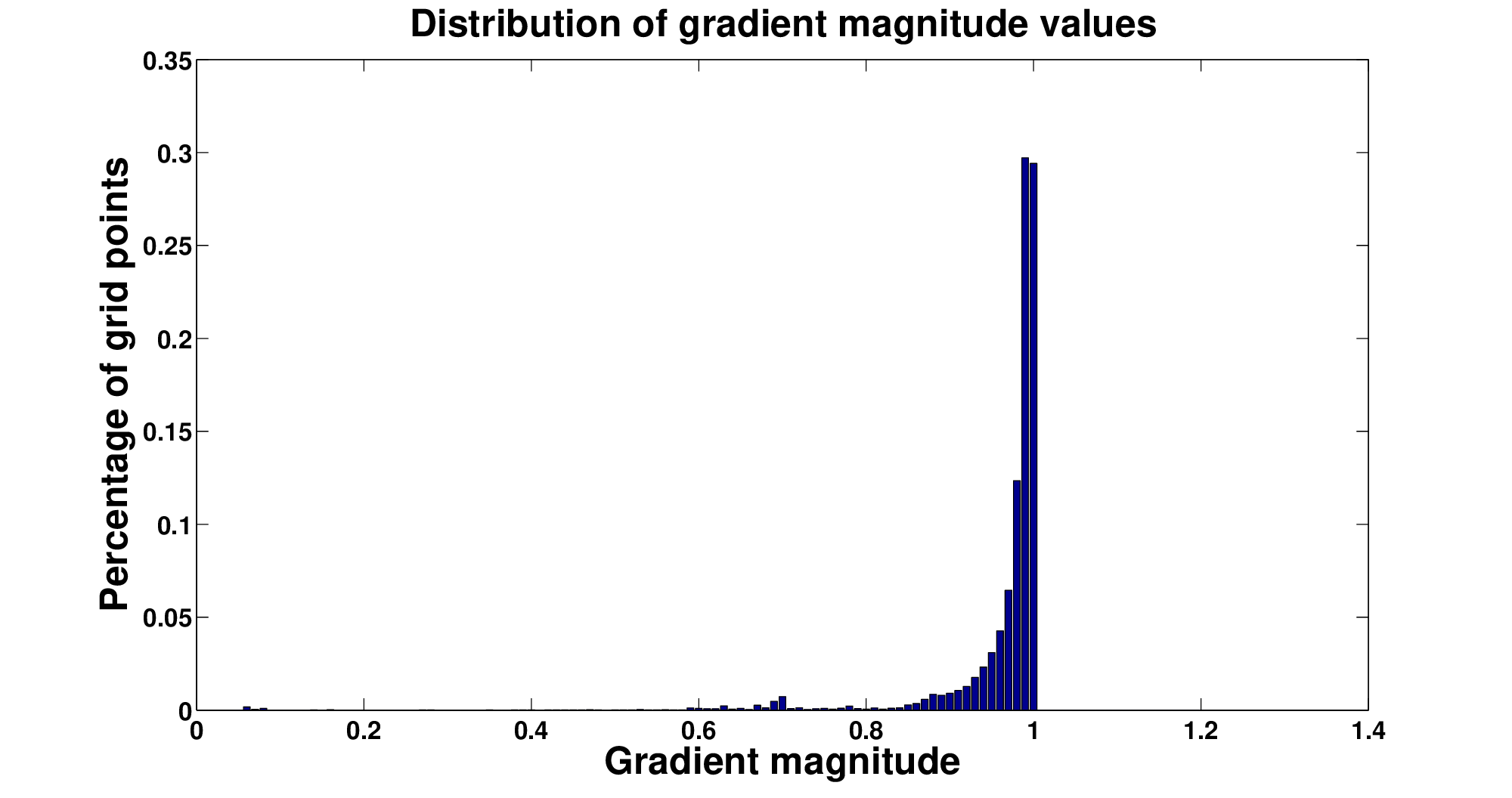}
\includegraphics[width=0.3\textwidth]{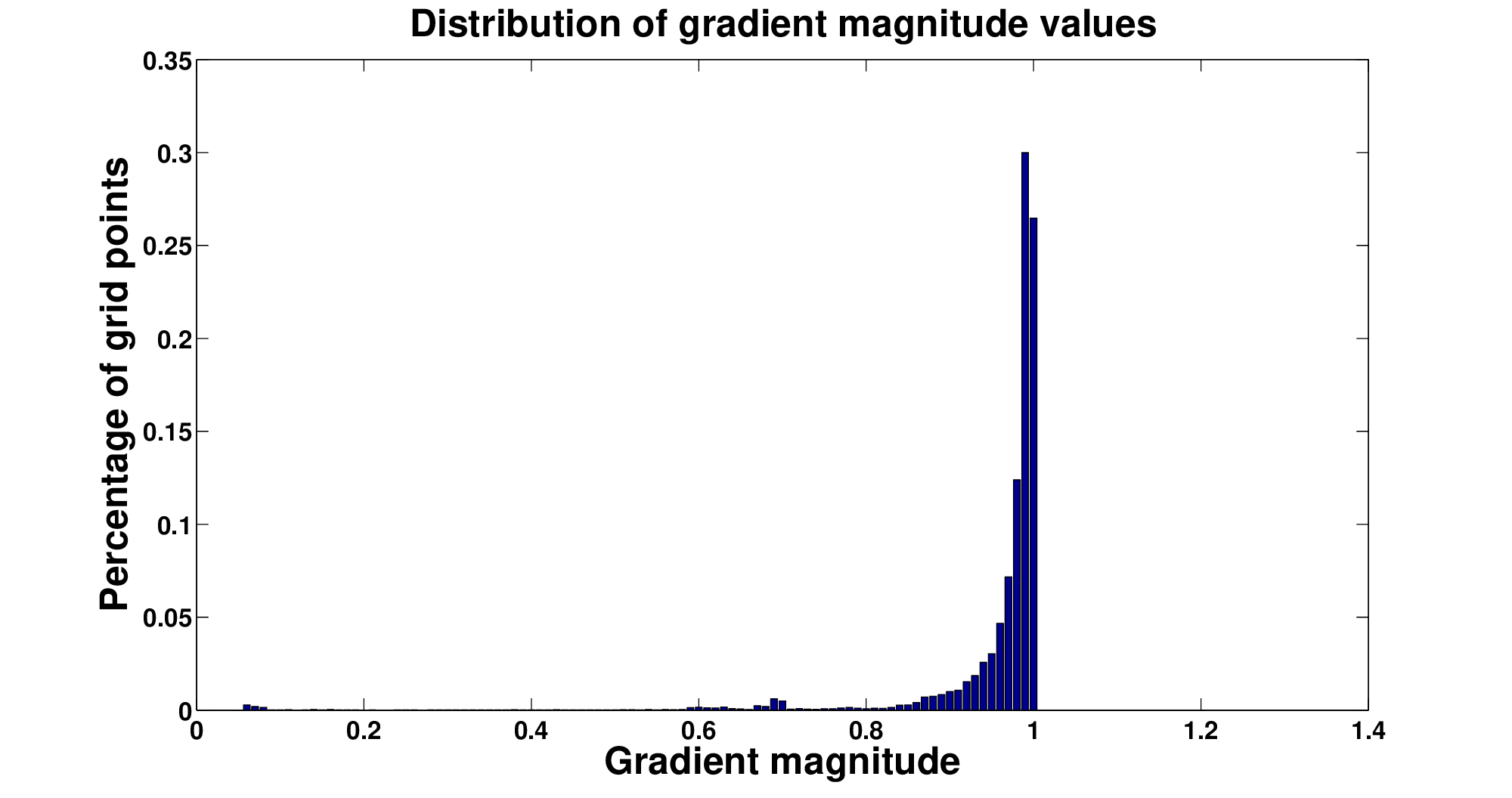}
\includegraphics[width=0.3\textwidth, height=0.22\textwidth]{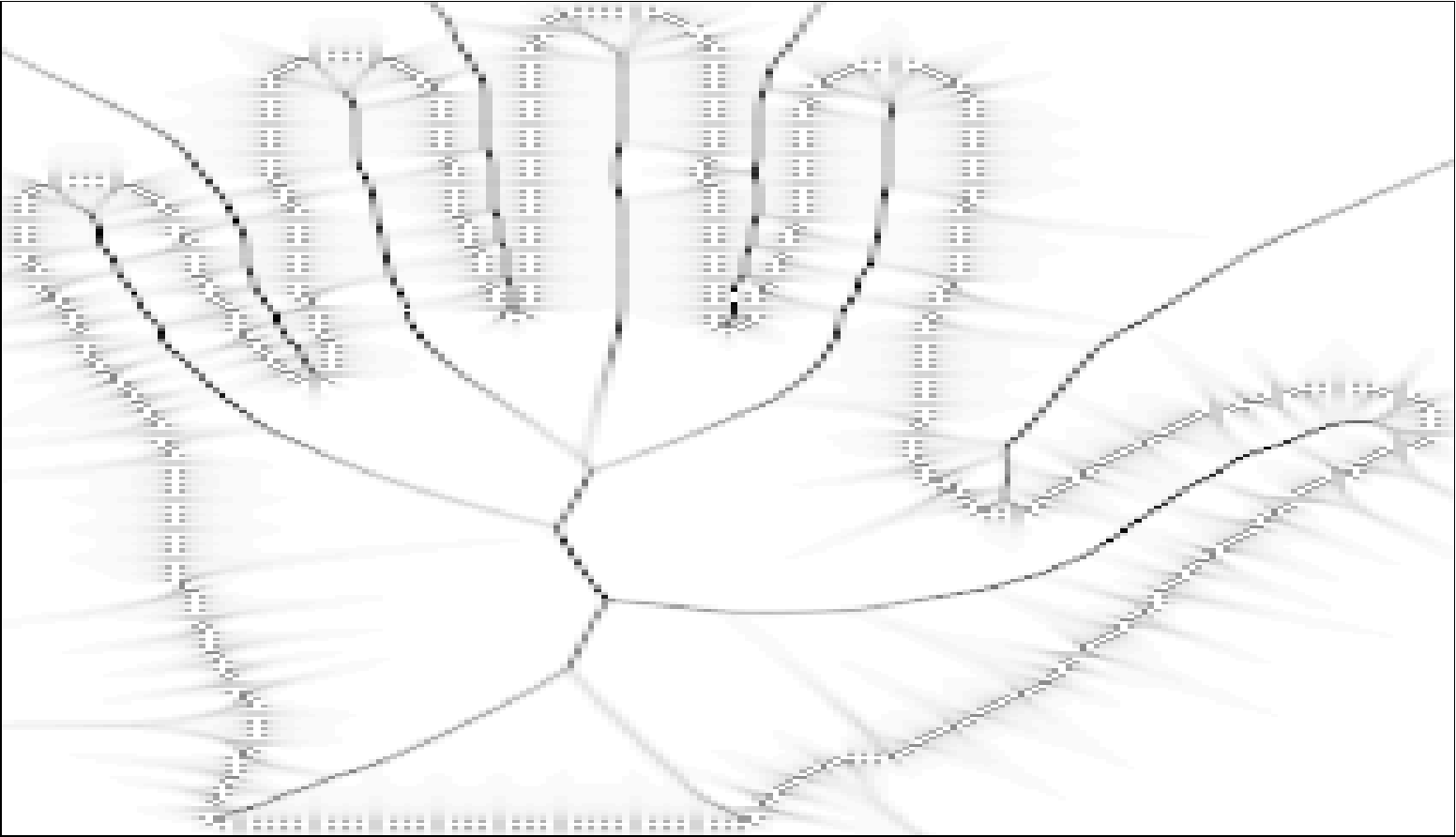}
\includegraphics[width=0.3\textwidth, height=0.22\textwidth]{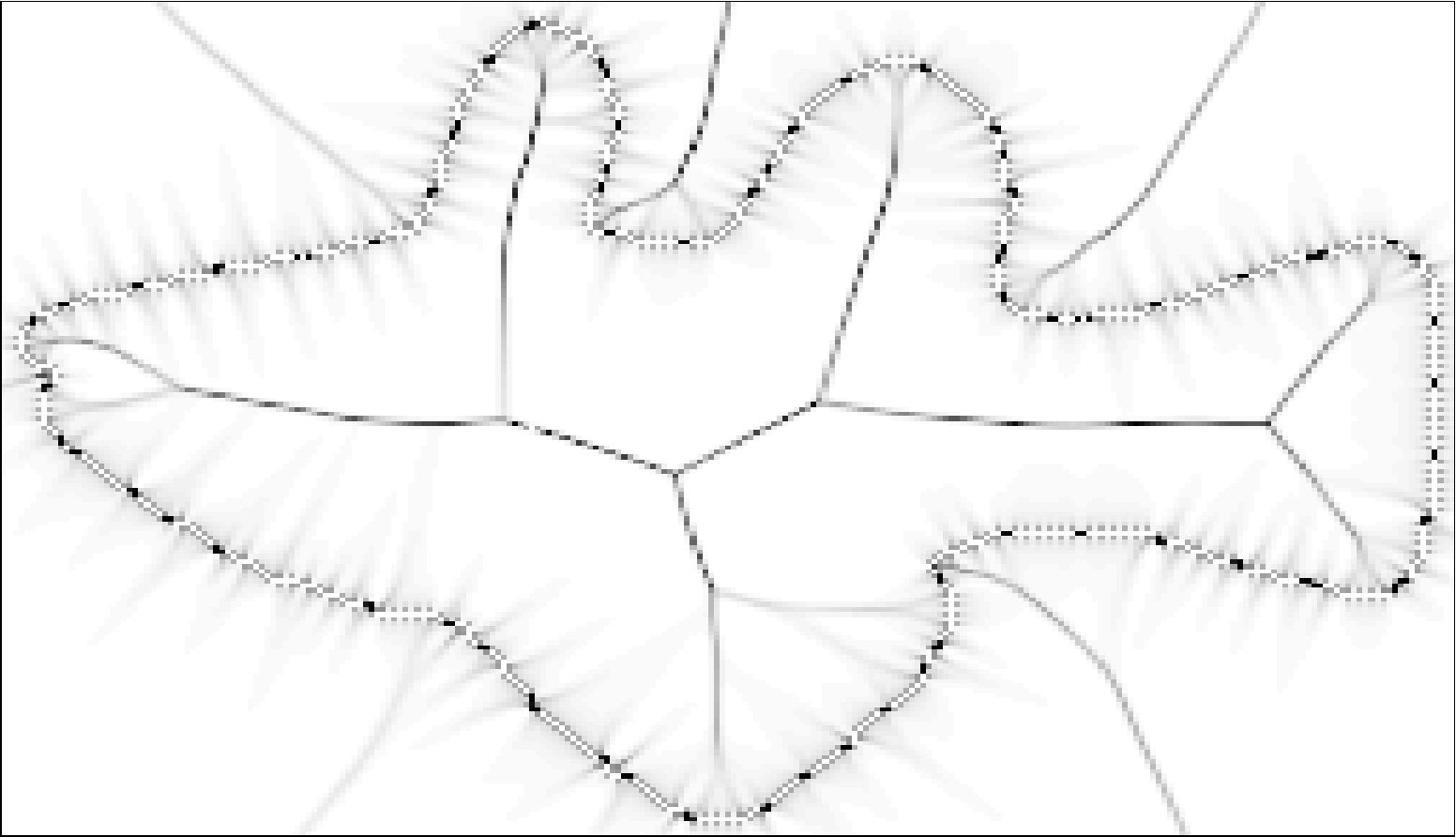}
\includegraphics[width=0.3\textwidth, height=0.22\textwidth]{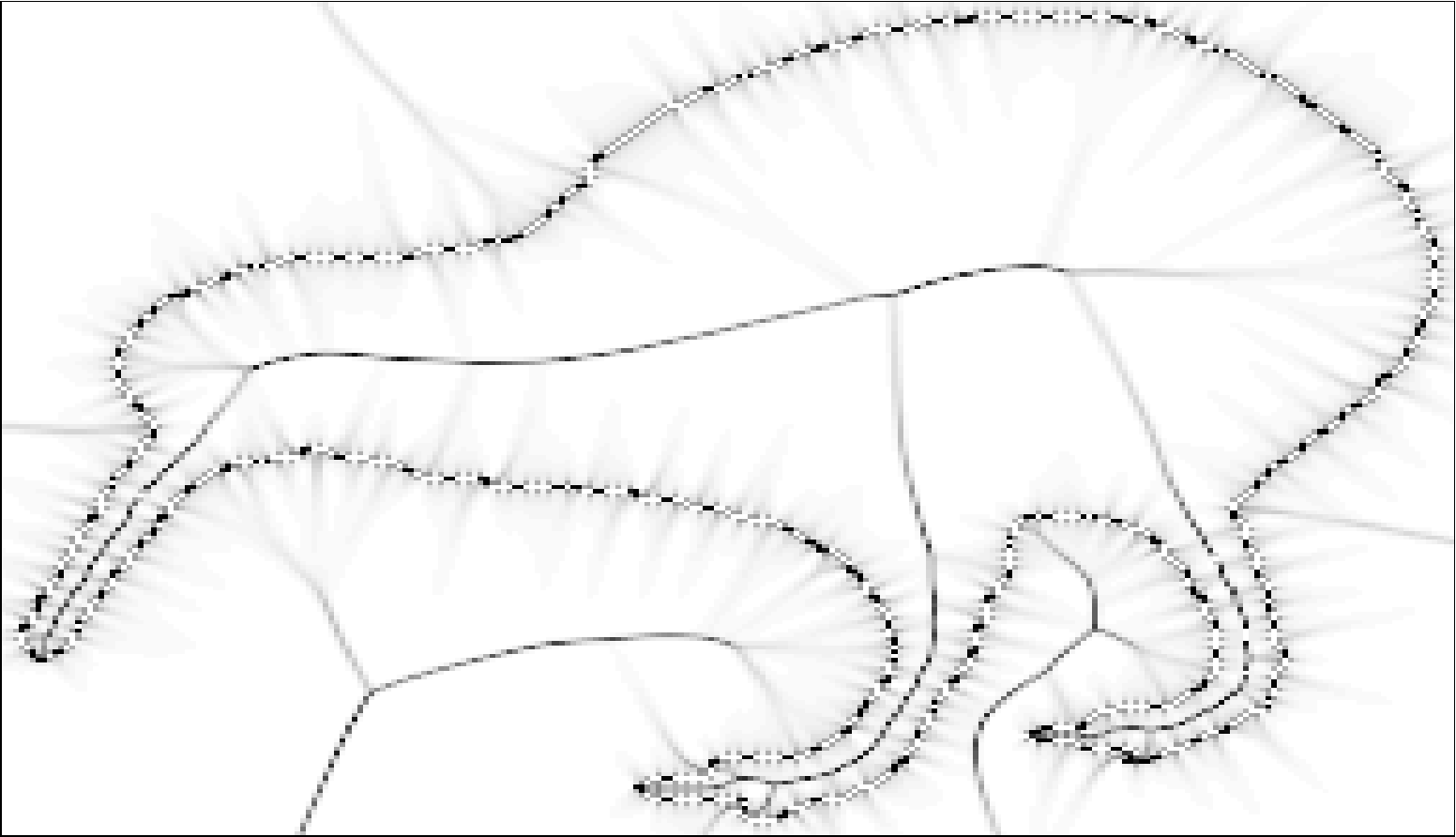}
\par\end{centering}
\caption{Gradient magnitude ($\|\nabla S\|$) (i) Top: Distribution, (ii) Bottom: Image plot
\label{fig:Gradmagnitude}}
\end{figure}
\par\end{center}
\vspace{-1cm}
\subsection{3D Experiments}
\vspace{0.3cm}
\noindent \textbf{Example 4:} We also compared our Euclidean distance function
fast convolution method with the fast sweeping method \cite{Zhao05} and the exact
Euclidean distance on the \emph{``Dragon''} point-set obtained from
the Stanford 3D Scanning Repository\footnote{This dataset is available at http://graphics.stanford.edu/data/3Dscanrep/.%
}.
The common grid was $-0.117\leq x\leq0.117$, $-0.086\leq y\leq0.086$ and $-0.047\leq z\leq0.047$ and the grid spacing equals $\frac{1}{2^8}$. The point cloud obtained from sampling the surface of the Dragon was rounded to the nearest grid points to generate the source locations. We ran our approach 
at $\tau = 0.0004$ and ran the fast sweeping method for 15 iterations which
is sufficient for the Gauss-Seidel iterations to converge. We then 
calculated the percentage error for both these techniques according to (\ref{eq:percentError}).
While the average percentage error for our approach when compared to the true distance 
function was just 1.306\%, the average percentage
error in the fast sweeping method was about 6.84\%. Our FFT-based approach does
not begin by discretizing the spatial differential operator as is
the case with the fast marching and fast sweeping methods and this
could help account for the increased accuracy. The iso-surface obtained by connecting the grid points at a distance
of $0.005$ from the point-set determined by the true Euclidean distance
function, fast convolution and fast sweeping are shown in Figure~\ref{fig:Isosurface}. 
The similarities between the plots provide anecdotal visual
evidence for the usefulness of our approach.
\begin{center}
\begin{figure}
\begin{centering}
\includegraphics[width=0.3\textwidth]{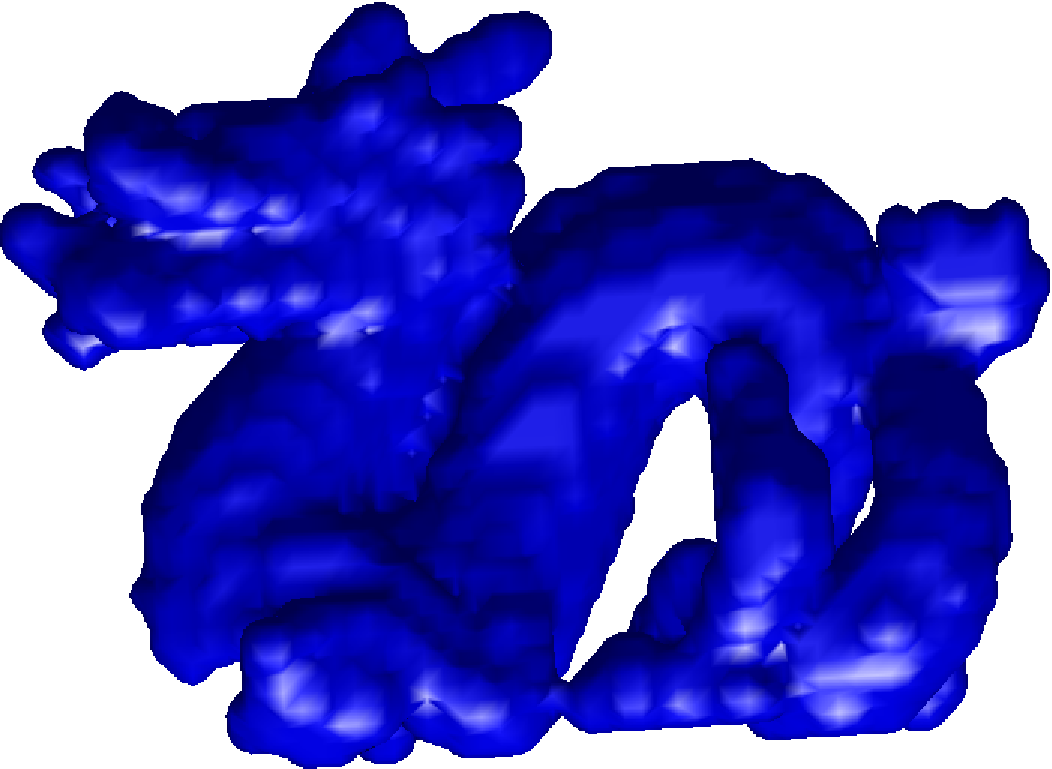} 
\includegraphics[width=0.3\textwidth]{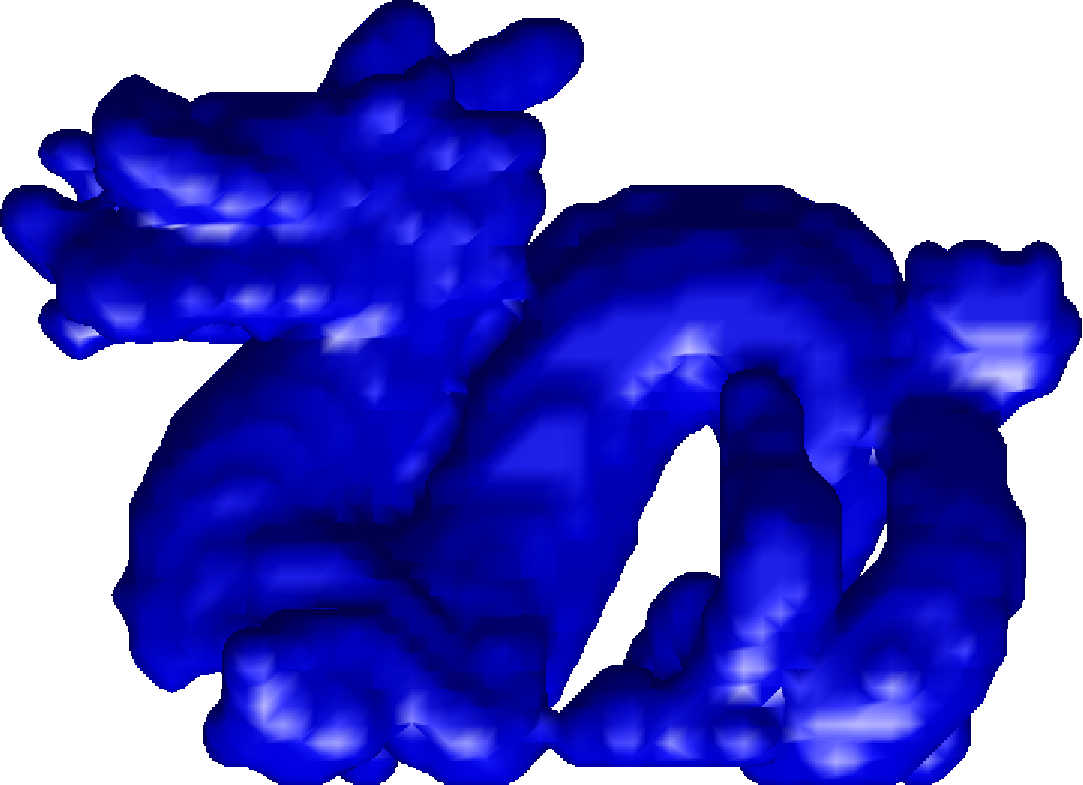}
\includegraphics[width=0.3\textwidth]{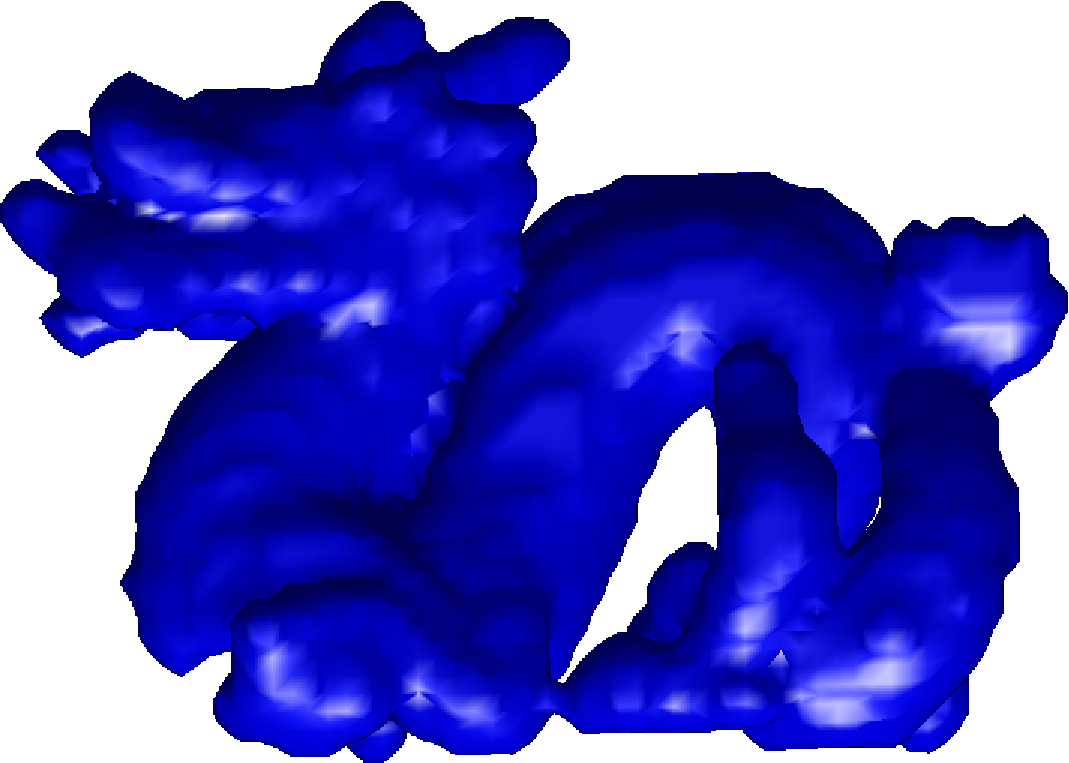} 
\par\end{centering}
\caption{Isosurfaces: (i) Left: Actual Euclidean distance function, (ii) Center:
Fast convolution and (iii) Right: Fast sweeping}
\label{fig:Isosurface} 
\end{figure}
\par\end{center}
\vspace{-0.5cm}
\noindent \textbf{Example 5:} We now demonstrate the efficacy of our 
fast convolution method for computing the topological degree. We considered a 3D grid,
$-0.125 \leq x \leq 0.125$, $-0.125 \leq y \leq 0.125$ and 
$-0.125 \leq z \leq 0.125$ at a grid width of $\frac{1}{2^8}$. 
Given a set of points sampled from the surface of a 3D 
object (and moved to the closest grid location), we triangulated the surface using built-in $\mbox{MATLAB}^{\textregistered}$ routines. 
We consider the incenter of each triangle to represent the data points $\{Y_k\}_{k=1}^K$. 
The normal $P_k$ for each triangle can be computed from the cross-product of the triangle vector edges. The direction of the normal vector
was determined by taking the dot product between the position vector $\vec{Y_k}$
and the normal vector $\vec{P_k}$. For negative dot products, $\vec{P_k}$ was 
negated to obtain a outward pointing normal vector. We then computed
the topological degree for all the $N$ grid locations \emph{simultaneously} 
in $O(N \log N)$ by running our fast convolution-based algorithm. Grid locations 
where the topological degree value equaled or exceeded $1$  were marked as 
inner points. Figure~\ref{fig:topdegree} 
shows the interior points for the three 3D objects---cube, sphere and
cylinder (left to right).
\begin{center}
\begin{figure}
\begin{centering}
\includegraphics[width=0.22\textwidth]{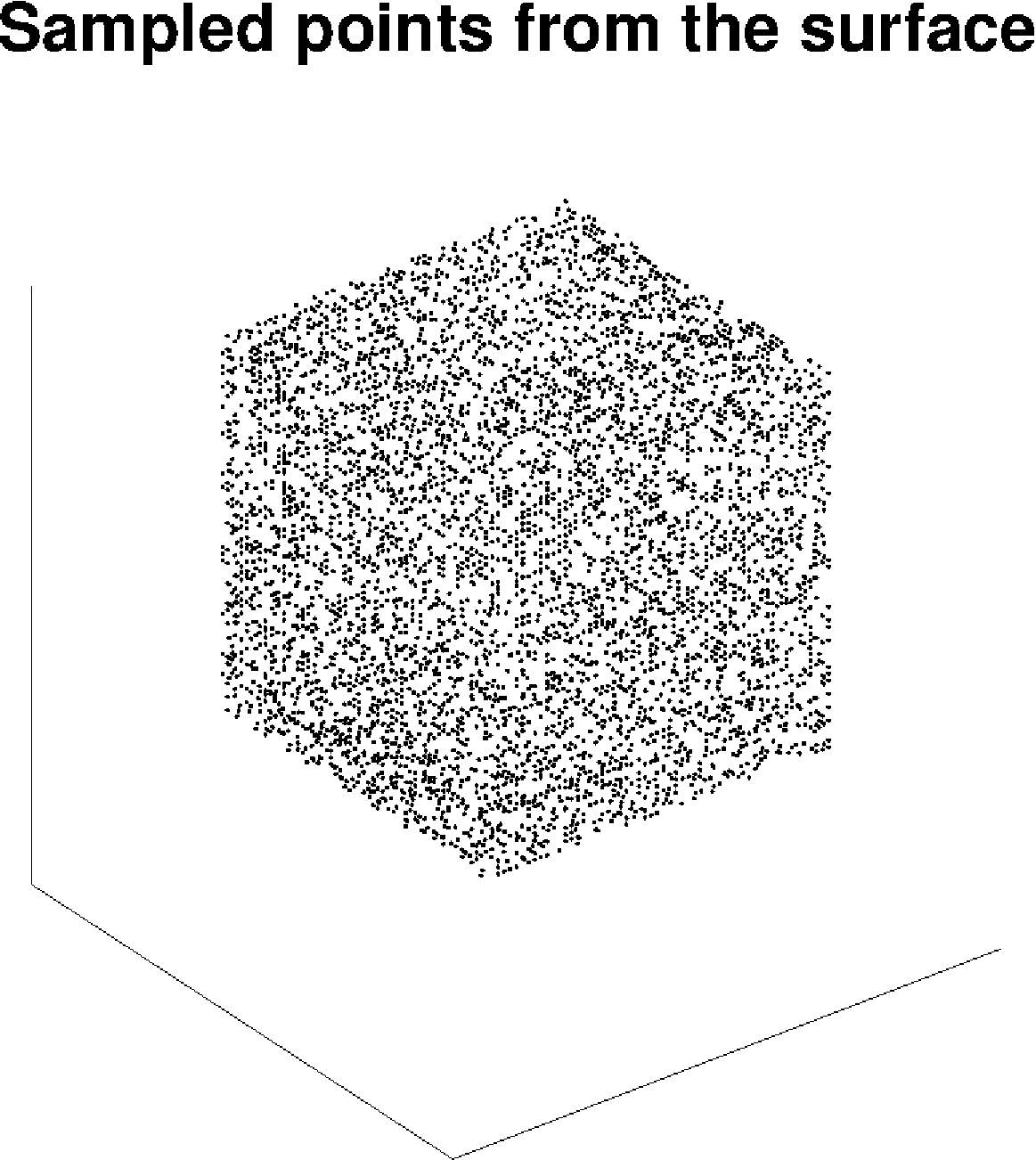}
\includegraphics[width=0.22\textwidth]{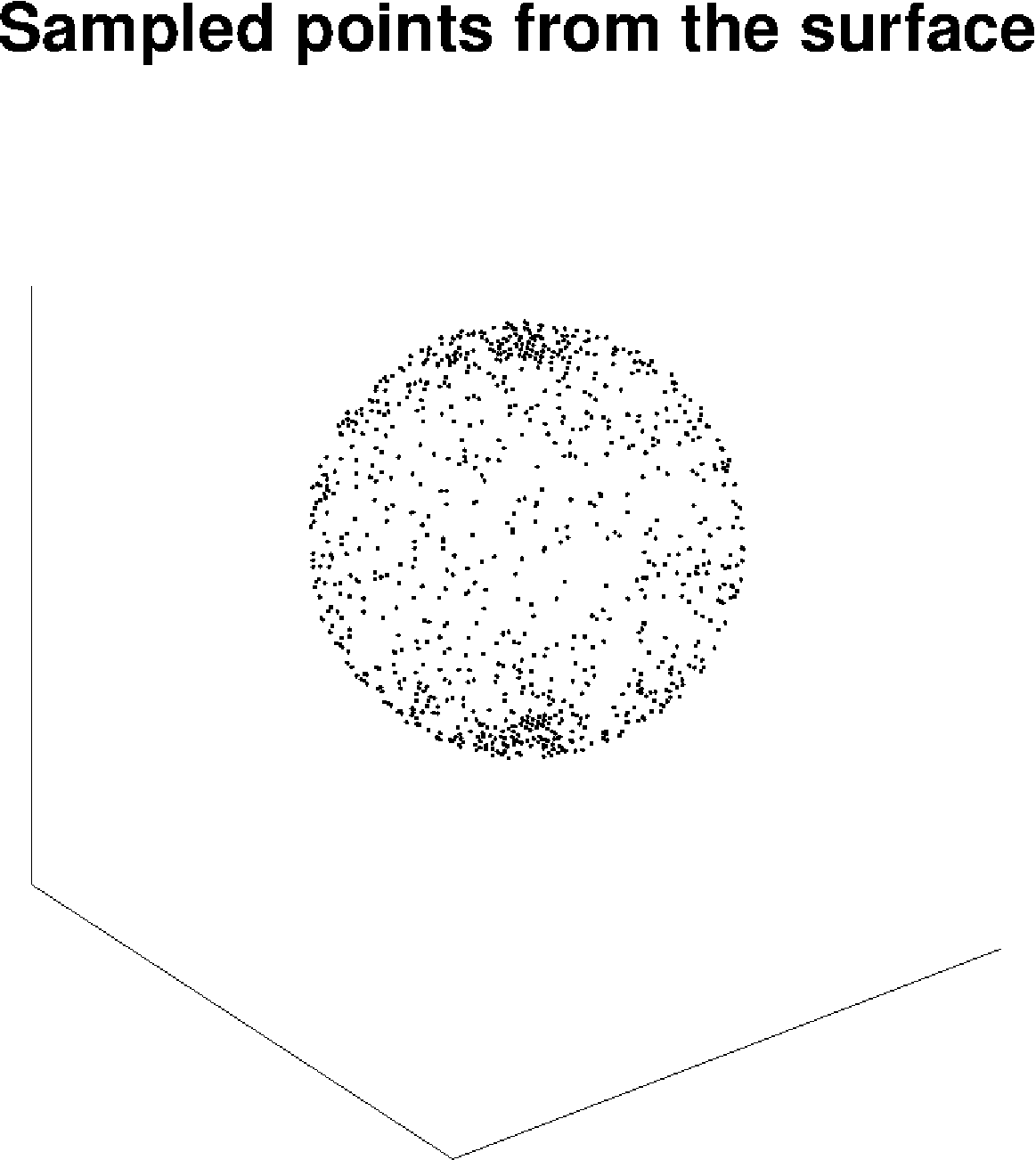}
\includegraphics[width=0.35\textwidth]{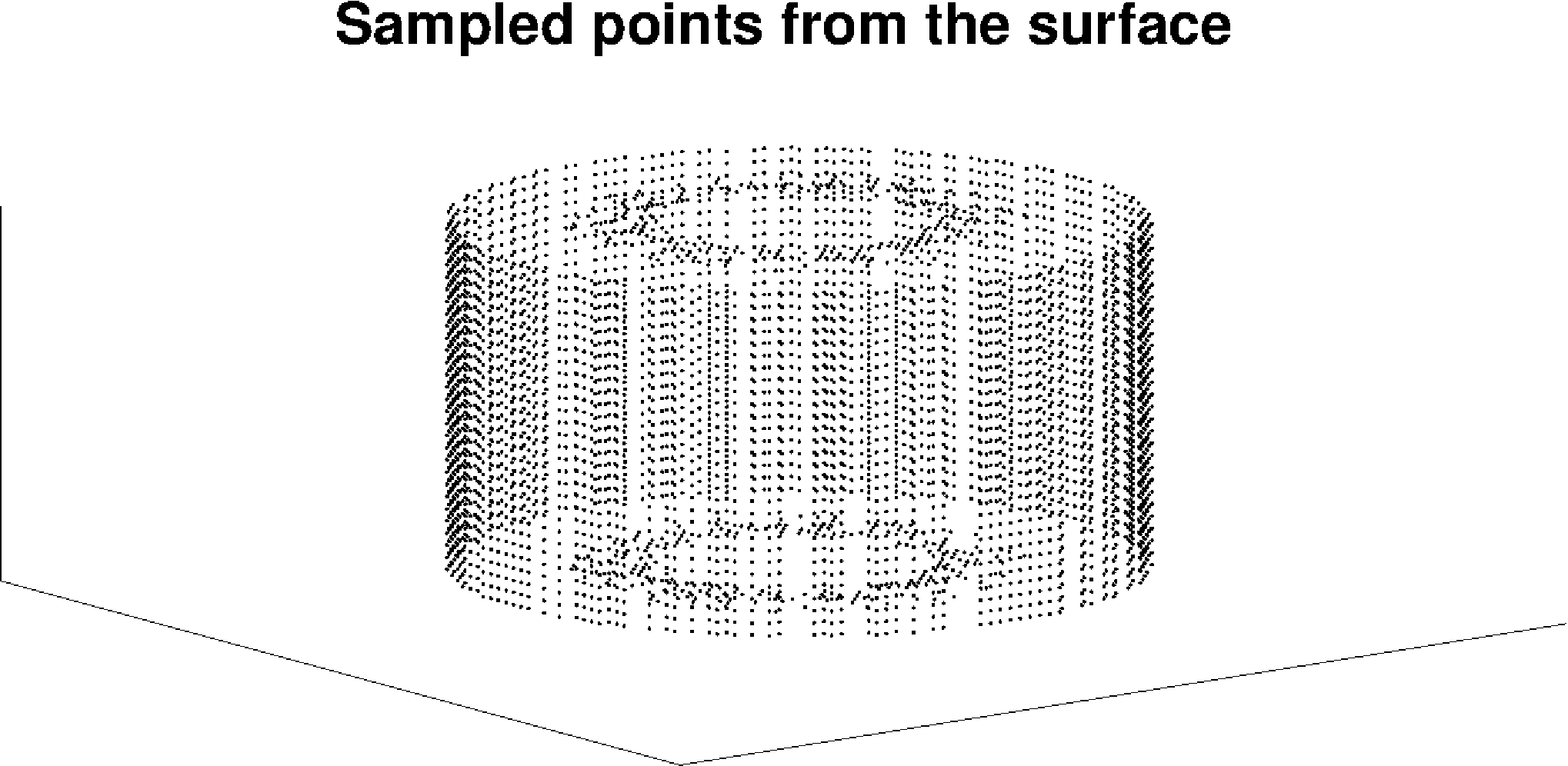}
\includegraphics[width=0.22\textwidth]{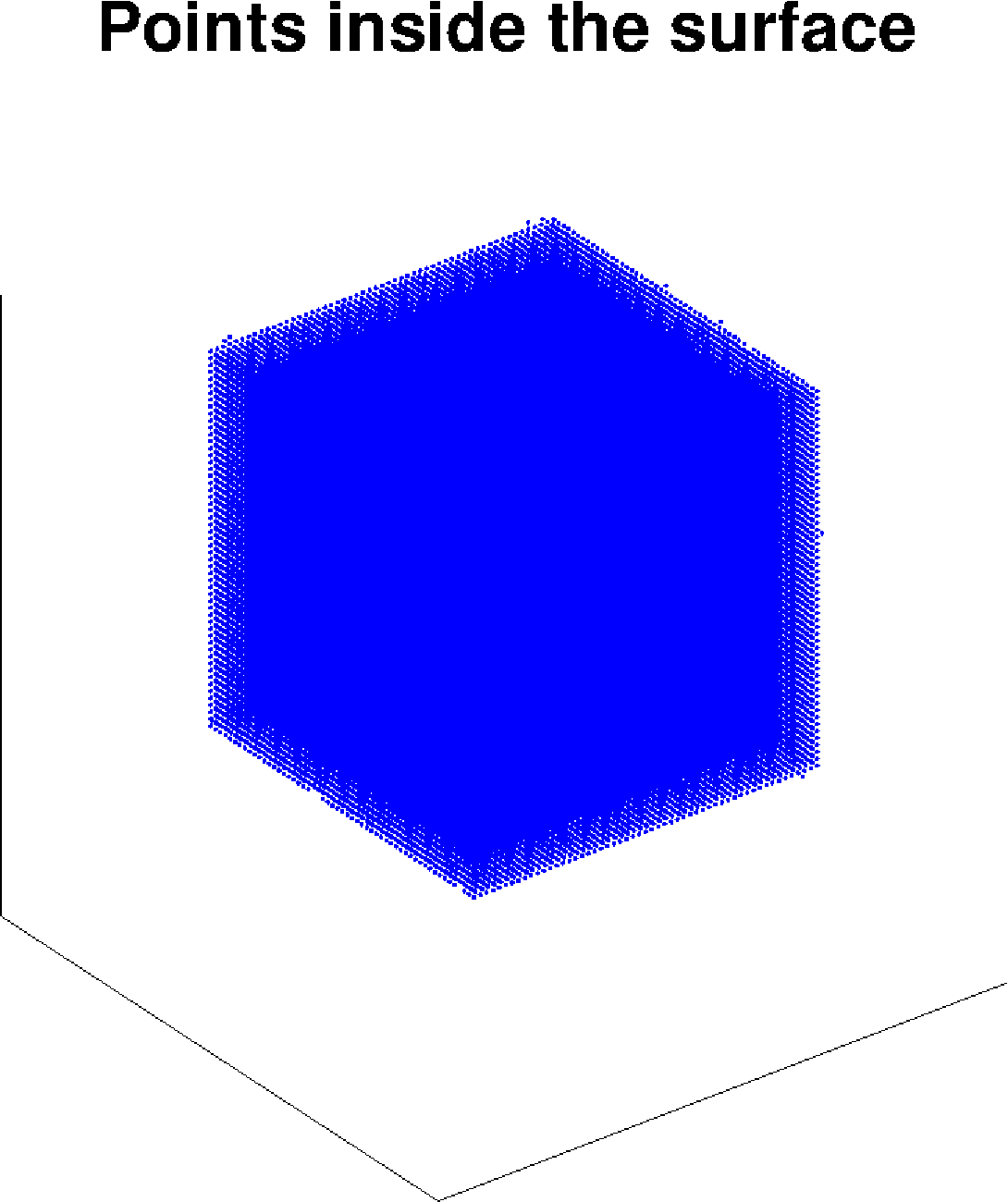}
\includegraphics[width=0.22\textwidth]{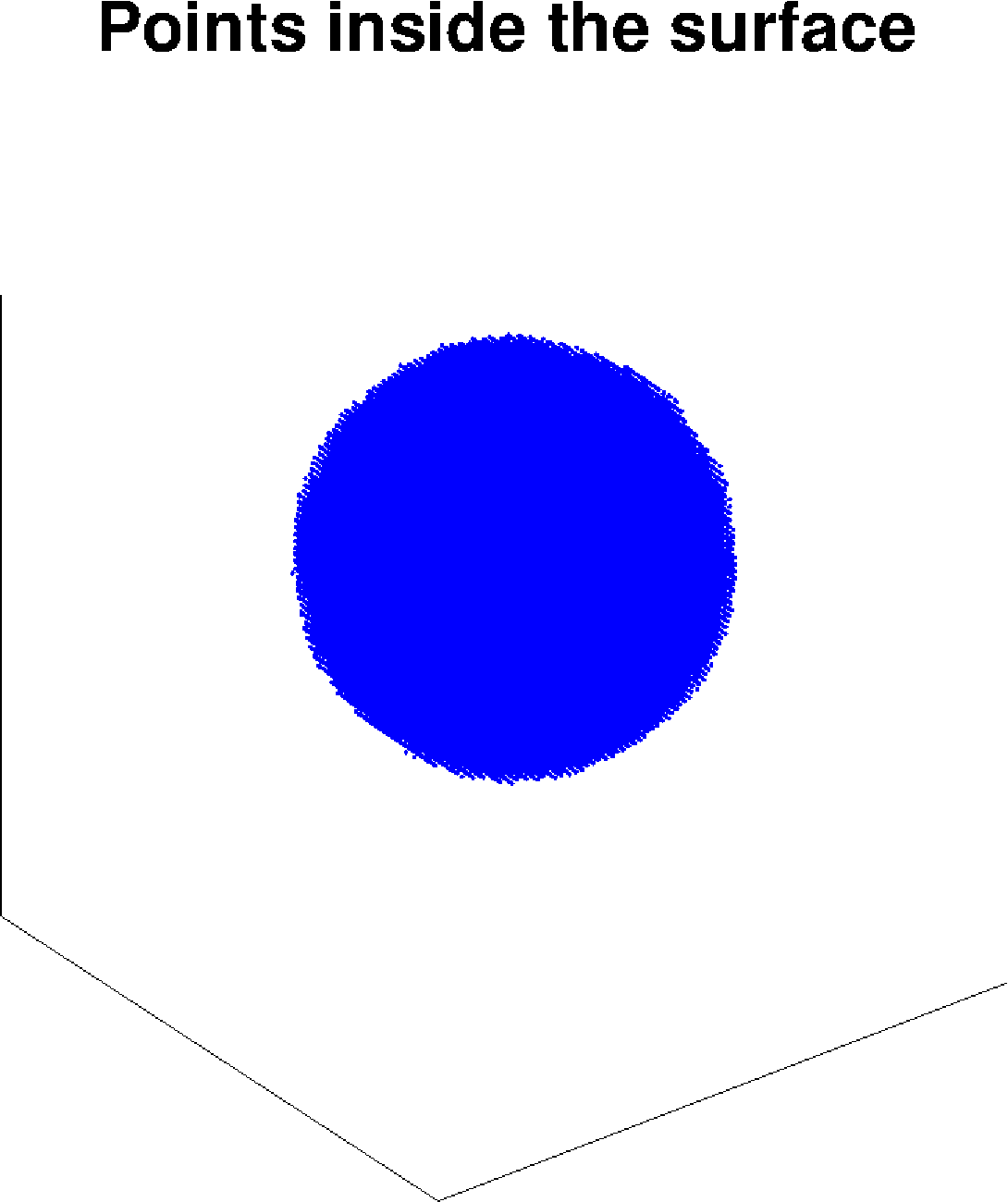}
\includegraphics[width=0.35\textwidth]{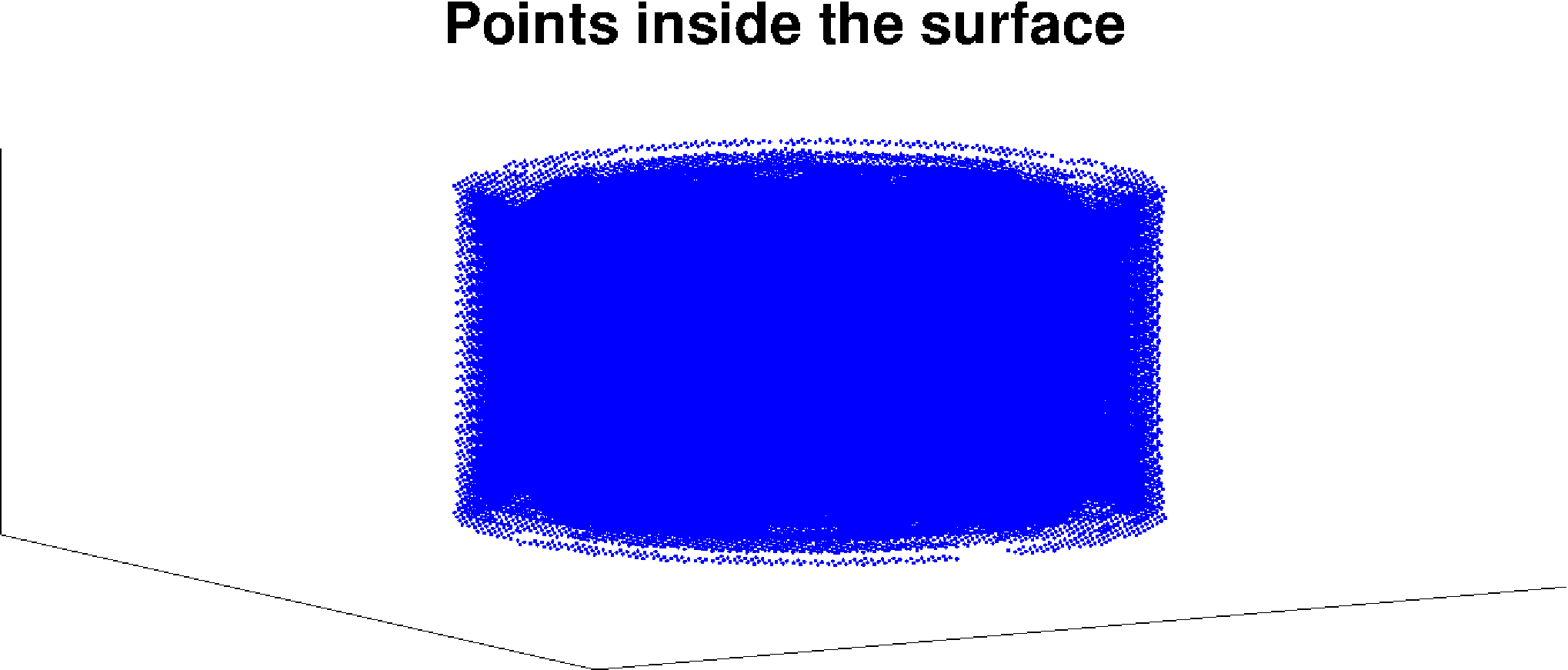}
\par\end{centering}
\caption{Topological Degree: (i) Top: Sampled points from the surface, 
(ii) Bottom: Grid points lying inside the surface (marked as blue)}
\label{fig:topdegree}
\end{figure}
\par\end{center}  
\vspace{-1cm}
\section{Conclusion}
\label{sec:Conclusion}
In this work we began with a variational formalism for the Euclidean distance function problem.
The variational problem with a free parameter $\tau$ balancing the data and
gradient terms led to an Euler-Lagrange equation---a linear differential
equation similar to a inhomogeneous, screened, Poisson equation. The intriguing aspect of our approach is that 
the non-linear Hamilton-Jacobi equation is embedded inside a \emph{linear}
differential equation and the solution is derived in the limiting case of $\tau \rightarrow 0$. We obtained the solution for the function $\phi$ satisfying the
Euler-Lagrange equation via a Green's function approach and later approximated it with a closed-form solution which does not require spatial discretization of the derivative operator (unlike the Hamilton-Jacobi solvers) and is computed in $O(N\log N)$ at the $N$ grid locations using a FFT-based convolution method. The Euclidean distance is then recovered by taking the negative logarithm of $\phi$. Since the scalar field $\phi$ is determined at a small but non-zero $\tau$, the obtained Euclidean distance function is an approximation. We derived analytic bounds for the error of the approximation for a given value of $\tau$ and provided proofs of convergence to the true distance function as $\tau \rightarrow 0$. The differentiability of our solution allowed us to express even the gradients of the distance function in closed-form, also written as convolutions. Finally, we demonstrated how our discrete convolution-based technique for computing the
winding number in 2D and the topological degree in 3D aid in determining the sign of the distance function. 

While Hamilton-Jacobi solvers have gone beyond the eikonal equation and 
regular grids---by providing efficient solutions even for the more general static 
Hamilton-Jacobi equation on irregular grids---in the current work we
restrict ourselves only to computing the Euclidean distance function on regular grids. 
In the future we would like to follow the pioneering works of the fast
marching and fast sweeping methods and try to extend our linear formalism to irregular grids and to
the more general eikonal equation. 

\appendix
\section{Convergence analysis for the distance function}
\label{sec:proof}
We provide the convergence of the distance function $S(X)$ to its true value $R(X)$ separately for each spatial dimension.\\
\noindent \textbf{1D:}
Using the expression for the 1D Green's function we solve for $\phi$ as,
 \begin{equation}
 \phi(X) = \frac{1}{2\tau^{\frac{3}{2}}} \sum_{k=1}^K \int_{X-Y_k-\frac{\tau}{2}}^{X-Y_k+\frac{\tau}{2}} \exp\left(-\frac{|Z|}{\tau}\right) dZ.
 \end{equation}
 From the relation~$\ref{SfromPhi}$ we get
 \begin{equation}
 S(X) = C_{\tau}-\tau \log \left( \sum_{k=1}^K \int_{\mathcal{B}_k^{\tau}(X)} \exp\left(-\frac{|Z|}{\tau}\right) dZ \right)
 \end{equation}
 where $C_{\tau} = \tau \log(2)+\frac{3}{2}\tau \log(\tau)$ and the integration region $\mathcal{B}_k^{\tau}(X)$ equals
 \begin{equation}
 \label{eq:IntergralRegion1D}
 \mathcal{B}_k^{\tau}(X) = \left[X-Y_k-\frac{\tau}{2},X-Y_k+\frac{\tau}{2}\right].
 \end{equation}
For each $k$ define
  \begin{equation}
 \alpha_k \equiv \left \{
                     \begin{array}{ll}
                     1 & \mbox{if $X>Y_k$};\\
                     -1 &\mbox{if $X<Y_k$}.
                     \end{array}
                     \right. 
 \end{equation}
Let $k_0$ denote the index of the source point closest to a given grid point
$X$, i.e, $R(X) =|X-Y_{k_0}| \leq |X-Y_k|, \forall k$. Then for sufficiently
small $\tau$ we have 
\begin{eqnarray}
S(X) &\leq& C_{\tau}- \tau \log \left\{ \tau \exp\left(-\frac{|X-Y_{k_0}+\alpha_{k_0} \frac{\tau}{2}|}{\tau}\right) \right\} \nonumber \\
&=&C_{\tau}-\tau \log (\tau)+\left |X-Y_{k_0}+\alpha_{k_0} \frac{\tau}{2} \right |
\end{eqnarray}
as $|X-Y_{k_0}+\alpha_{k_0} \frac{\tau}{2}| \geq |Z|$ for $Z \in \mathcal{B}_{k_0}^{\tau}$.

On the other hand, notice that for each $k$, $|X-Y_{k}-\alpha_{k}
\frac{\tau}{2}| \leq |Z|, \forall Z \in \mathcal{B}_k^{\tau}$. Hence for small
values of $\tau$ we also arrive at the inequality 
\begin{eqnarray}
S(X) &\geq&C_{\tau}-\tau \log \left\{ \tau \sum_{k=1}^K \exp\left(-\frac{|X-Y_{k}-\alpha_{k} \frac{\tau}{2}|}{\tau}\right) \right\} \nonumber \\
&\geq&C_{\tau}-\tau \log (\tau) - \tau \log \left \{ K \exp\left(-\frac{|X-Y_{k_0}-\alpha_{k_0} \frac{\tau}{2}|}{\tau}\right) \right\} \nonumber \\
&=& C_{\tau}-\tau \log (\tau) -\tau \log(K) +  \left |X-Y_{k_0}-\alpha_{k_0} \frac{\tau}{2} \right |.
\end{eqnarray}
In order to see why the second step in the above relation holds, consider the
two scenarios: (i) $X$ is not a point on the Voronoi boundary, and (ii) $X$
lies on the Voronoi boundary. If $X$ doesn't lie on the Voronoi boundary, then
exists a neighborhood $N_p (X)$ around $X$ such that $\forall Y \in  N_p (X)$,
$|Y-Y_{k_0}| < |Y-Y_k|, \forall k$. Since $|X-\alpha_{k_0} \frac{\tau}{2}| \in
N_p (X)$ for sufficiently small values of $\tau$, the aforementioned relation
is true. On the flip side, if $X$ is a point on the Voronoi boundary, the
closest source point $Y_{k}$ is not uniquely defined. However we can
unambiguously choose a closest source point $Y_{k_0}$ and a $\tau_0$ such that
for $\tau \in (0,\tau_0]$, $|X-\alpha_{k_0} \tau-Y_{k_0}| < |X-\alpha_{k}
  \tau-Y_{k}|, \forall k$. These observations buttress the above inequality.  

 Since $C_\tau$, $\tau \log \tau$ and $\tau \log K$ approach zero as $\tau
 \rightarrow 0$, we obtained the desired result namely, $\lim_{\tau
   \rightarrow 0} S(X) = |X-Y_{k_0}| = R(X)$.\\\\ 
 \noindent \textbf{2D:}
Based on the expression for the 2D Green's function and the relation~\ref{SfromPhi}, we get
 \begin{equation}
 S(X) = C_{\tau}-\tau \log \left( \sum_{k=1}^K \int_{\mathcal{B}_k^{\tau}(X)} K_0\left(\frac{\|Z\|}{\tau}\right) dZ \right),
 \end{equation}
where $C_{\tau} = \tau \log (2 \pi) + 3 \tau \log \tau$ and the integral region $\mathcal{B}_k^{\tau}(X)$ equals
\begin{equation}
 \label{eq:IntergralRegion2D}
\mathcal{B}_k^{\tau}(X) = \left[ x-x_k-\frac{\tau}{2}, x-x_k+\frac{\tau}{2}\right] \times  \left[y-y_k-\frac{\tau}{2}, y-y_k+\frac{\tau}{2}\right] .
\end{equation}
Defining
 \begin{equation}
 \alpha_k \equiv \left \{
                     \begin{array}{ll}
                     1 & \mbox{if $x>x_k$};\\
                     -1 &\mbox{if $x<x_k$}
                     \end{array}
                     \right. \mbox{and} \hspace{5pt}
 \beta_k \equiv \left \{
                     \begin{array}{ll}
                     1 & \mbox{if $y>y_k$};\\
                     -1 &\mbox{if $y<y_k$}
                     \end{array}
                     \right .
\end{equation}
for each $k$ and closely following the arguments illustrated for the 1D case, we have
\begin{equation}
S(X) \leq C_{\tau}-\tau \log (\tau)-\tau \log \left\{K_0\left(\frac{\|X_{1\tau}-Y_{k_0}\|}{\tau}\right) \right\} 
\end{equation}
for small values of $\tau$, where $X_{1\tau} = \left(x+\alpha_k
\frac{\tau}{2},y+\beta_k \frac{\tau}{2} \right)$. As considered above, $k_0$
denotes the index of the source point ($Y_{k_0}$) closest to $X$. 
Similar to the 1D case, note that $\|X_{1\tau}-Y_{k_0}\|\geq \|Z\|$ for $Z \in
\mathcal{B}_{k_0}$ where $Y_{k_0}$ is the closest source-point to $X$. Using the relation $K_0(z)\geq \frac{\exp(-z)}{\sqrt{z}}$ when $z\geq0.5$ we also get
\begin{equation}
S(X) \leq  C_{\tau}-\tau \log (\tau) +\tau \log \left\{\sqrt{\frac{\|X_{1\tau}-Y_{k_0}\|}{\tau}}\right\}+\|X_{1\tau}-Y_{k_0}\|
 \end{equation}
 as $\tau \rightarrow 0$. If we let $X_{2\tau}= \left(x-\alpha_k \frac{\tau}{2},y-\beta_k \frac{\tau}{2} \right)$, then similar to the 1D case we arrive at the inequality,
\begin{eqnarray}
S(X)&\geq& C_{\tau}-\tau \log \left\{ \tau \sum_{k=1}^K K_0\left(\frac{\|X_{2\tau}-Y_{k}\|}{\tau}\right) \right\} \nonumber \\
&\geq&C_{\tau}-\tau \log (\tau) - \tau \log \left \{ K K_0\left(\frac{\|X_{2\tau}-Y_{k_0}\|}{\tau}\right) \right\} \nonumber \\
&=& C_{\tau}-\tau \log (\tau) -\tau \log(K) - \tau \log \left \{K_0\left(\frac{\|X_{2\tau}-Y_{k_0}\|}{\tau}\right) \right\}.
\end{eqnarray}
As $K_{0}(z)\leq\exp(-z)$ when $z\geq1.5$, it follows that
 \begin{equation}
S(X) \geq  C_{\tau}-\tau \log (\tau) -\tau \log(K)  + \|X_{2\tau}-Y_{k_0}\|
 \end{equation}
 for small values  of $\tau$. Since $X_{1\tau}, X_{2\tau}$ approach $X$ as
 $\tau \rightarrow 0$ and the rest of the terms tend to zero, we obtain the
 desired result namely, 
 $\lim_{\tau \rightarrow 0} S(X) = \|X-Y_{k_0}\| = R(X)$.\\\\
 \noindent \textbf{3D:}
 By exactly following the line of argument delineated for the 1D and the 2D
 case where we bound $S(X)$ above and below by functions which 
converge to the true Euclidean distance function $R(X)$ as $\tau \rightarrow
0$, we can prove that $\lim_{\tau \rightarrow 0} S(X)= R(X)$. 
 
\bibliographystyle{elsarticle-num}
\bibliography{EuclideanDist}

\begin{thebibliography}{10}
\expandafter\ifx\csname url\endcsname\relax
  \def\url#1{\texttt{#1}}\fi
\expandafter\ifx\csname urlprefix\endcsname\relax\def\urlprefix{URL }\fi
\expandafter\ifx\csname href\endcsname\relax
  \def\href#1#2{#2} \def\path#1{#1}\fi

\bibitem{Osher02}
S.~J. Osher, R.~P. Fedkiw, Level set methods and dynamic implicit surfaces,
  Springer-Verlag, New York, NY, 2003.

\bibitem{Aggarwal87}
A.~Aggarwal, L.~Guibas, J.~Saxe, P.~Shor, A linear time algorithm for computing
  the voronoi diagram of a convex polygon, in: STOC, ACM, 1987, pp. 39--45.

\bibitem{deBerg08}
M.~D. Berg, O.~Cheong, M.~V. Kreveld, M.~Overmars, Computational geometry:
  {A}lgorithms and applications, Springer-Verlag, New York, NY, 2008.

\bibitem{Sethian96}
J.~A. Sethian, A fast marching level set method for monotonically advancing
  fronts, Proc. Nat. Acad. Sci. 93~(4) (1996) 1591--1595.

\bibitem{Yatziv06}
L.~Yatziv, A.~Bartesaghi, G.~Sapiro, O({N}) implementation of the fast marching
  algorithm, J. Comp. Phys. 212~(2) (2006) 393--399.

\bibitem{Zhao05}
H.~K. Zhao, A fast sweeping method for eikonal equations, Math. Comp. 74~(250)
  (2005) 603--627.

\bibitem{Kao02}
C.~Y. Kao, S.~J. Osher, Y.~H. Tsai, Fast sweeping methods for static
  {H}amilton-{J}acobi equations, SIAM J. Num. Anal. 42~(6) (2004) 2612--2632.

\bibitem{Qian07}
J.~Qian, Y.~T. Zhang, H.~K. Zhao, Fast sweeping methods for eikonal equations
  on triangular meshes, SIAM J. Num. Anal. 45~(1) (2007) 83--107.

\bibitem{Kao08}
C.~Y. Kao, S.~J. Osher, J.~Qian, Legendre-transform-based fast sweeping methods
  for static {H}amilton-{J}acobi equations on triangulated meshes, J. Comp.
  Phys. 227~(24) (2008) 10209--10225.

\bibitem{Siddiqi99}
K.~Siddiqi, A.~Tannenbaum, S.~Zucker, A {H}amiltonian approach to the eikonal
  equation, in: EMMCVPR, Vol. 1654 of LNCS, Springer, 1999, pp. 1--13.

\bibitem{Tsai02}
Y.~H.~R. Tsai, Rapid and accurate computation of the distance function using
  grids, J. Comp. Phys. 178 (2002) 175--195.

\bibitem{Danielsson80}
P.~E. Danielsson, Euclidean distance mapping, Comp. Graph. Image Proc. 14~(3)
  (1980) 227--248.

\bibitem{Gurumoorthy09}
K.~S. Gurumoorthy, A.~Rangarajan, A {S}chr\"odinger equation for the fast
  computation of approximate {E}uclidean distance functions, in: SSVM, Vol.
  5567 of LNCS, Springer, 2009, pp. 100--111.

\bibitem{Rangarajan09}
A.~Rangarajan, K.~S. Gurumoorthy, A {S}chr\"odinger wave equation approach to
  the eikonal equation: {A}pplication to image analysis, in: EMMCVPR, Vol. 5681
  of LNCS, Springer, 2009, pp. 140--153.

\bibitem{Sethi12}
M.~Sethi, A.~Rangarajan, K.~S. Gurumoorthy, The {S}chr\"odinger {D}istance
  {T}ransform {(SDT)} for point-sets and curves, in: CVPR, IEEE Computer
  Society, 2012, pp. 198--205.

\bibitem{Bracewell99}
R.~N. Bracewell, The {F}ourier transform and its applications, 3rd Edition,
  McGraw-Hill, New York, NY, 1999.

\bibitem{Wahba90}
G.~Wahba, Spline models for observational data, Vol.~59 of CBMS-NSF Regional
  Conference Series in Applied Mathematics, SIAM, Philadelphia, PA, 1990.

\bibitem{Abramowitz64}
M.~Abramowitz, I.~A. Stegun, Handbook of mathematical functions with formulas,
  graphs and mathematical tables, Dover, New York, NY, 1964.

\bibitem{Cooley65}
J.~W. Cooley, J.~W. Tukey, An algorithm for the machine calculation of complex
  {F}ourier series, Math. Comp. 19~(90) (1965) 297--301.

\bibitem{Fousse07}
L.~Fousse, G.~Hanrot, V.~Lef\'evre, P.~P\'elissier, P.~Zimmermann, {MPFR}: A
  multiple-precision binary floating-point library with correct rounding, ACM
  Trans. Math. Softw. 33 (2007) 1--15.

\bibitem{GMP}
G.~Torbj\"orn, \emph{et al.}, {GNU} multiple precision arithmetic library 5.0.1
  (June 2010).

\bibitem{Advanpix}
Multiprecision computing toolbox for {MATLAB} ({A}dvanpix {LLC.}) (2014).

\bibitem{Gray97}
A.~Gray, Modern differential geometry of curves and surfaces with mathematica,
  2nd Edition, CRC Press, Boca Raton, FL, 1997.

\bibitem{Aberth98}
O.~Aberth, Precise numerical methods using {C}++, Academic Press, San Diego,
  CA, 1998.

\end{thebibliography}
\end{document}